\documentclass[lettersize,journal]{IEEEtran}
\usepackage{amsmath,amsfonts}
\usepackage{array}\usepackage[caption=false,font=normalsize,labelfont=sf,textfont=sf]{subfig}
\usepackage{textcomp}
\usepackage{stfloats}
\usepackage{url}
\usepackage{verbatim}
\usepackage{graphicx}
\usepackage{cite}
\usepackage{multirow}
\usepackage{multicol}
\usepackage{booktabs}
\usepackage[table,xcdraw,x11names]{xcolor}
\usepackage{colortbl}
\usepackage{bbding}
\usepackage{algorithm}
\usepackage{algpseudocode}
\usepackage[percent]{overpic}
\usepackage{diagbox}
\usepackage{caption}
\usepackage[colorlinks,
            linkcolor=red,
            anchorcolor=blue,
            citecolor=green
            ]{hyperref}

\definecolor{ztfgray}{rgb}{0.85, 0.85, 0.85}

\begin{document}

\title{S2AFormer: Strip Self-Attention for Efficient Vision Transformer}
% \title{FSAViT: Factorized Self-Attention for Efficient Vision Transformer}

\author{
Guoan Xu,
Wenfeng Huang,
Wenjing Jia,~\IEEEmembership{Member,~IEEE}, 
Jiamao Li,~\IEEEmembership{Member,~IEEE}, 
Guangwei Gao,~\IEEEmembership{Senior Member,~IEEE}, 
Guo-Jun Qi,~\IEEEmembership{Fellow,~IEEE,}
        % <-this % stops a space
%\thanks{This paper was produced by the IEEE Publication Technology Group. They are in Piscataway, NJ.}% <-this % stops a space
\thanks{This work was supported in part by the foundation of Key Laboratory of Artificial Intelligence of Ministry of Education under Grant AI202404.~\textit{(Corresponding authors: Wenjing Jia, Guangwei Gao.)}} 

\thanks{Guoan Xu, Wenfeng Huang, and Wenjing Jia are with the Faculty of Engineering and Information Technology, University of Technology Sydney, Sydney, NSW 2007, Australia (e-mail: xga\_njupt@163.com, huang-wenfeng@outlook.com, and Wenjing.Jia@uts.edu.au).}

\thanks{Jiamao Li is with the Bionic Vision System Laboratory, State Key Laboratory of Transducer Technology, Shanghai Institute of Microsystem and Information Technology, Chinese Academy of Sciences, Shanghai 200050, China (e-mail:jmli@mail.sim.ac.cn).}

\thanks{Guangwei Gao is with the PCA Lab, Key Lab of Intelligent Perception and Systems for High-Dimensional Information of Ministry of Education, School of Computer Science and Engineering, Nanjing University of Science and Technology, Nanjing 210094, China, and also with the Key Laboratory of Artificial Intelligence, Ministry of Education, Shanghai 200240, China (e-mail: csggao@gmail.com).}

\thanks{Guo-Jun Qi is with the Research Center for Industries of the Future and the School of Engineering, Westlake University, Hangzhou 310024, China, and also with OPPO Research, Seattle, WA 98101 USA (e-mail: guojunq@gmail.com).}
}

\markboth{IEEE Transactions on Image Processing}%
{Shell \MakeLowercase{\textit{et al.}}: A Sample Article Using IEEEtran.cls for IEEE Journals}


\maketitle

\begin{abstract}

The Vision Transformer (ViT) has achieved remarkable success in computer vision due to its powerful token mixer, which effectively captures global dependencies among all tokens. However, the quadratic complexity of standard self-attention with respect to the number of tokens severely hampers its computational efficiency in practical deployment. Although recent hybrid approaches have sought to combine the strengths of convolutions and self-attention to improve the performance–efficiency trade-off, the costly pairwise token interactions and heavy matrix operations in conventional self-attention remain a critical bottleneck. To overcome this limitation, we introduce S2AFormer, an efficient Vision Transformer architecture built around a novel Strip Self-Attention (SSA) mechanism. Our design incorporates lightweight yet effective Hybrid Perception Blocks (HPBs) that seamlessly fuse the local inductive biases of CNNs with the global modeling capability of Transformer-style attention. The core innovation of SSA lies in simultaneously reducing the spatial resolution of the key ($K$) and value ($V$) tensors while compressing the channel dimension of the query ($Q$) and key ($K$) tensors. This joint spatial-and-channel compression dramatically lowers computational cost without sacrificing representational power, achieving an excellent balance between accuracy and efficiency. We extensively evaluate S2AFormer on a wide range of vision tasks, including image classification (ImageNet-1K), semantic segmentation (ADE20K), and object detection/instance segmentation (COCO). Experimental results consistently show that S2AFormer delivers substantial accuracy improvements together with superior inference speed and throughput across both GPU and non-GPU platforms, establishing it as a highly competitive solution in the landscape of efficient Vision Transformers.
\end{abstract}


\section{Introduction}
Vision Transformers (ViTs)~\cite{carion2020end,dosovitskiy2020image} have emerged as a notable competitor to CNNs, demonstrating a superior ability to capture long-range interactions between image patches. While Convolutional Neural Networks (CNNs) limit interactions to local regions through shared kernels, ViTs divide the input image into patches and use self-attention (SA) to update token features, enabling global interactions. Consequently, Transformer-based architectures have gained increasing attention in computer vision, achieving remarkable results across various tasks such as image classification~\cite{liu2021swin,touvron2021training}, object detection~\cite{carion2020end,maaz2022class}, and semantic segmentation~\cite{cheng2022masked,xie2021segformer}. 

% \begin{figure}[t]
%     \centering
%     % 使用 overpic 在图像上叠加表格
%     \begin{overpic}[width=\columnwidth]{modelpics/a.pdf} % 主图
%         % 在右下角叠加表格
%         \put(35, 8){ % 调整 (x, y) 坐标以定位表格
%             \resizebox{0.6\columnwidth}{!}{ % 调整表格大小
%                 \begin{tabular}[b]{l|ccc}
%                     \hline
%                     \footnotesize
%                     Model & Top-1 Accuracy & \# Parameters & \# MMAC \\
%                     \hline
%                     EfficientViT-M5~\cite{liu2023efficientvit} & 77.1\% & 12.4M & 522 \\
%                     EdgeViT~\cite{chen2022edgevit} &77.5\% & 6.7M &1100\\
%                     EMO-5M~\cite{zhang2023rethinking} & 78.4\% & 5.1M & 903 \\
%                     \rowcolor{gray!10}\bf S2AFormer-XS & \bf 78.9\% & \bf 6.5M & \bf 786 \\
%                     FastViT-T12~\cite{vasu2023fastvit}& 79.1\% & 6.8M & 1400\\
%                     SHViT-S4~\cite{yun2024shvit} & 82.0\% & 16.5M & 3973 \\
%                     PVTv2-B2~\cite{wang2021pvtv2} & 82.0\% & 25.4M & 4000 \\
%                     InceptionNeXt-T~\cite{yu2024inceptionnext} & 82.3\% & 28.0M & 4200 \\
%                     \rowcolor{gray!10}\bf S2AFormer-M & \bf 82.3\% & \bf 24.9M & \bf 4120 \\ \hline
%                 \end{tabular}
%             }
%         }
%     \end{overpic}
%     \vspace{-0.2cm}
%     \caption{\small{Comparison of our proposed S2AFormer model with SOTA methods (Top-1 accuracy v.s. MACs on ImageNet-1k~\cite{deng2009imagenet}).}}
%     \label{fig:imagenet_acc_flops}
%     \vspace{-2mm}
% \end{figure}
\begin{figure}[t]
    \centerline{\includegraphics[width=9cm]{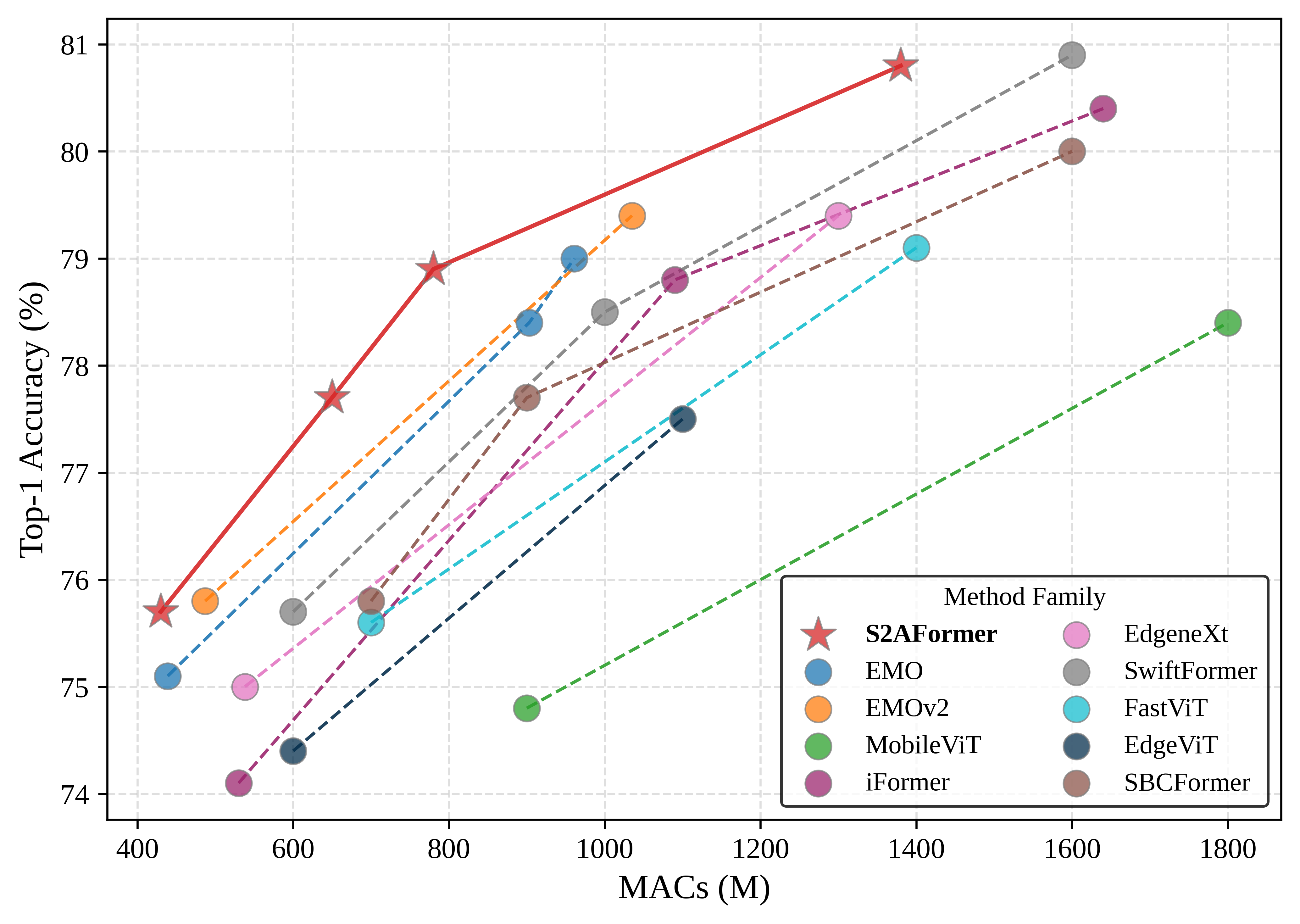}}
    \caption{\small{Comparison of our proposed S2AFormer model with SOTA methods, including EMOv2~\cite{zhang2025emov2}, SwiftFormer~\cite{shaker2023swiftformer}, and iFormer~\cite{zheng2025iformer}. (Top-1 accuracy v.s. MACs on ImageNet-1k~\cite{deng2009imagenet}). In theory, the optimal performance is in the upper-left of the plot, which means higher top-1 accuracy and fewer MACs.}}
    \label{fig:imagenet_acc_flops}
    \vspace{-2mm}
\end{figure}

However, the quadratic complexity %concerning the sequence length 
in the token mixer makes self-attention inefficient in terms of parameters and computational load. For images of size $h \times w$, the complexity $O(h^2w^2)$ limits the applicability of ViTs in high-resolution tasks (such as detection and segmentation), %, rendering them 
unsuitable for real-time applications and edge devices due to their significant computational overhead. % and slow processing. 

To address computational overhead, various optimized self-attention strategies have been introduced. Methods such as~\cite{liu2021swin,dong2022cswin} have been developed to limit the self-attention operation to local windows, rather than to the entire feature map. However, this approach compromises global information interaction between patches. 
Another approach involves token pruning or merging to selectively compute only the most informative tokens and eliminate uninformative tokens. For instance, Dynamic-ViT~\cite{rao2021dynamicvit} reduces %the token count 
the number of tokens by eliminating redundant tokens, while EViT~\cite{liang2022not} merges redundant tokens into a single token. DVT~\cite{wang2021not} adopts a flexible and dynamic patch-splitting strategy (\textit{e.g.}, $4 \times 4$, $7 \times 7$, etc.) tailored to the complexity of each image, moving away from the standard $14 \times 14$ approach. MG-ViT~\cite{zhang2024mg} emphasizes the need for different regions of an image to receive attention at varying levels of granularity, offering a more nuanced and effective representation. While reducing tokens decreases computation,  it presents challenges when dealing with complex, object-rich images, as it becomes increasingly difficult to determine which tokens should be retained.

Some approaches~\cite{wang2021pyramid,wang2021pvtv2,ren2023sg} aim to streamline the global sequence length by aggregating tokens across the entire key-value feature space, implementing a coarse-grained global attention mechanism. For instance, the Pyramid Vision Transformer (PVT)~\cite{wang2021pyramid} uses a large kernel with significant strides to achieve uniform token aggregation, generating a consistent coarse representation across the feature map. Efficient-ViT~\cite{xie2023efficient} introduces grouped attention heads to capture multi-scale features while distributing the spatial dimensions of each token, thereby effectively balancing the computational load. Similarly, SG-Former~\cite{ren2023sg} combines the windowing mechanism of the Swin Transformer~\cite{liu2021swin} with the spatial dimensionality reduction strategy of PVT~\cite{wang2021pyramid}, allowing for the extraction of information at different scales. Notably, it further optimizes efficiency by reducing the sequence length of $Q$, significantly decreasing computational complexity. 
%However, 
As pointed out in MetaSeg~\cite{kang2024metaseg}, beyond spatial redundancy, significant redundancy also exists in the channel dimension. 
By leveraging this channel redundancy, %observation, 
the computational complexity can be further reduced. 
%Meanwhile, 
While ViTs excel at modeling long-range dependencies, they tend to struggle with local feature representation compared to CNNs, a limitation that is often overlooked in the aforementioned architectures. 

\begin{figure}[t]
    \centering
    % 第一行
    \begin{minipage}[b]{0.23\textwidth}
        \centering
        \includegraphics[width=0.95\linewidth]{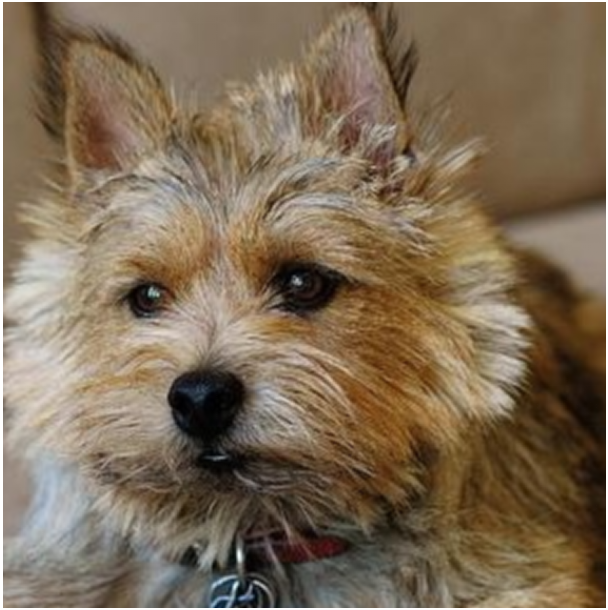}
        \small (a) Input image
    \end{minipage}
    \hfill
    \begin{minipage}[b]{0.23\textwidth}
        \centering
        \includegraphics[width=0.95\linewidth]{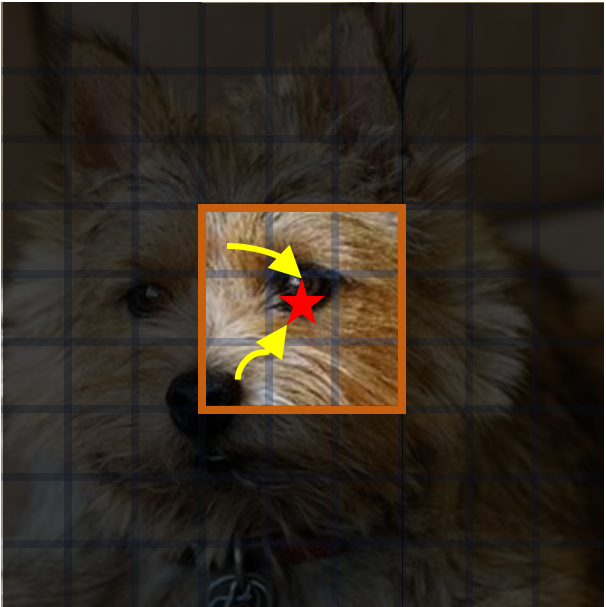}
        \small (b) Local perception
    \end{minipage}
    \vspace{1em}
    % 第二行
    \begin{minipage}[b]{0.23\textwidth}
        \centering
        \includegraphics[width=0.95\linewidth]{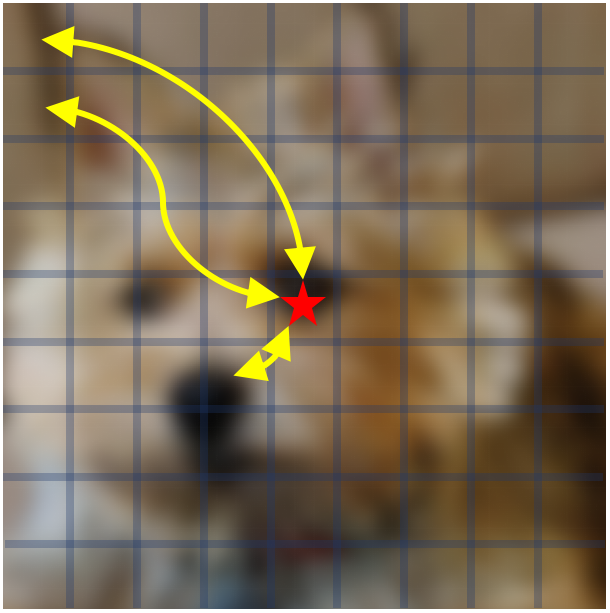}
        \small (c) Global self-attention
    \end{minipage}
    \hfill
    \begin{minipage}[b]{0.23\textwidth}
        \centering
        \includegraphics[width=0.95\linewidth]{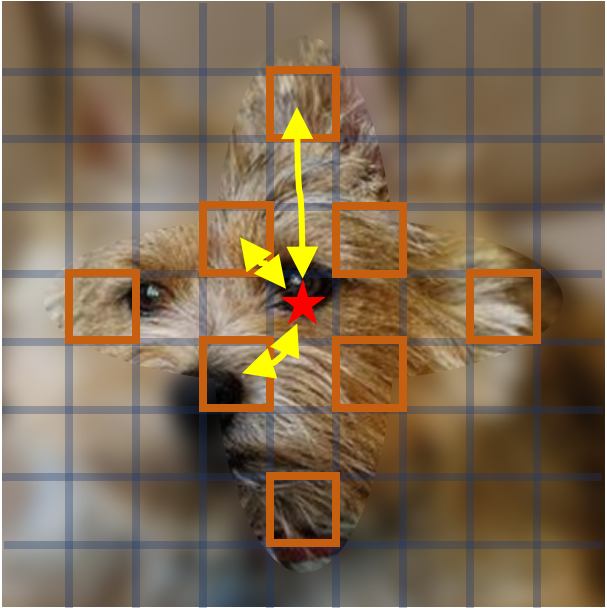}
        \small (d) \textbf{SSA + Local interaction}
    \end{minipage}
    \vspace{-2mm}
    \caption{Visualization of perception fields in different strategies. Red stars represent the current positions, and the black areas represent the regions that the current position cannot perceive. Convolutional networks cannot model long-range contexts. Global self-attention establishes global receptive fields at the cost of quadratic computational complexity. SSA integrated with local interaction prioritizes semantically salient regions globally while suppressing redundant spatial correlations. }
    \label{percaption}
    \vspace{-2mm}
\end{figure}

A comparative analysis of receptive field properties and cross-region information exchange mechanisms of different architectures is illustrated in Fig.~\ref{percaption}. 
Convolutional networks excel in local sensitivity due to their inductive bias toward spatial locality (via small kernels), enabling fine-grained detail extraction but struggling to model long-range contextual relationships. Conversely, vanilla Transformers establish global receptive fields through all-to-all token interactions, achieving holistic scene understanding at the cost of quadratic computational complexity and suboptimal local structural awareness. 

%Building on the above insights and analysis, 
In this paper, we propose %introduce 
an efficient % novel 
Transformer architecture, named \textit{S2AFormer}. Our goal is to address the primary challenge of preserving both local sensitivity and global receptive fields while reducing heavy computational load. 
To achieve this, we first design simple yet effective Hybrid Perception Blocks (HPBs) in the backbone network. 
Specifically, we propose a novel Strip Self-Attention (SSA) module that first applies spatial convolutions to the $Key$ and $Value$ features, thereby extracting more distilled feature representations. Subsequently, the $Query$ and $Key$ undergo channel-wise compression, significantly improving memory efficiency and reducing computational complexity from $O(n^2 \cdot C)$ to $O((\frac{n}{r} )^{2} \cdot h)$, where $r$ is the reduction rate and $h$ is the number of attention heads.

Meanwhile, rotations and translations are common data augmentations in vision tasks, and a well-designed model should ideally produce consistent predictions under such transformations, demonstrating a desirable degree of geometric invariance. However, Transformers relying on absolute positional encodings often introduce patch-specific biases that anchor predictions to fixed spatial coordinates, thereby undermining this invariance. Moreover, purely global attention mixing tends to neglect fine-grained local neighborhood structures and precise boundary details within patches. To address these limitations, we incorporate a dedicated Local Interaction Module (LIM) into HPB. This module effectively complements global receptive fields with localized processing, significantly enhancing boundary fidelity and robustness to rotations and translations. 
%Thus, our method reconciles these complementary strengths through dynamic sparse attention – prioritizing semantically salient regions while filtering redundant spatial correlations – thereby attaining human vision-aligned efficiency without compromising multi-scale representational capacity.
The LIM is implemented using only lightweight depthwise separable and grouped convolutions, ensuring minimal additional computational overhead. 
Furthermore, we integrate SENet~\cite{hu2018squeeze} to dynamically reweight features,  enabling the model to focus on semantically salient information within the local receptive field. 
A comprehensive series of rigorous experiments confirms that our S2AFormer architecture either rivals or surpasses the most advanced state-of-the-art backbones (as shown in Fig.~\ref{fig:imagenet_acc_flops}).

% A comparative analysis of receptive field properties and cross-region information exchange mechanisms of different architectures is illustrated in Fig.~\ref{percaption}. Convolutional networks excel in local sensitivity due to their inductive bias toward spatial locality (via small kernels), enabling fine-grained detail extraction but struggling to model long-range contextual relationships. Conversely, vanilla Transformers establish global receptive fields through all-to-all token interactions, achieving holistic scene understanding at the cost of quadratic computational complexity and suboptimal local structural awareness. 
% Our method reconciles these complementary strengths through dynamic sparse attention – prioritizing semantically salient regions while filtering redundant spatial correlations – thereby attaining human vision-aligned efficiency without compromising multi-scale representational capacity.

The key contributions of this paper are summarized as:
\begin{enumerate}
\item 
We propose a novel token-mixer mechanism, termed SSA,  which enables token interaction in a lightweight manner through both spatial and channel-wise compression, effectively reducing computational cost and achieving faster inference. 
\item 
We introduce HPB, a simple yet effective design that enhances the global self-attention with local perceptual capabilities. 
By incorporating the LIM module, HPB improves the model’s ability to capture fine-grained visual details, producing more comprehensive and expressive feature representations.
\item 
We pretrain our backbone on ImageNet-1K~\cite{deng2009imagenet} and evaluate its generalization capability on multiple downstream tasks, including object detection, instance segmentation, and semantic segmentation, all of which 
demonstrate strong 
%achieve strong 
performance, providing a solid pretrained model for future research.
\end{enumerate}

\begin{table*}[!t]
\caption{Key properties observed from successful backbones that benefit model design.}
\begin{center}
\begin{tabular}{l|ccccc}
\toprule
\diagbox{Methods}{Properties} & Global dependency & Local perception & Low computation & High performance & Fast inference \\
\midrule
% EdgeViT~\cite{chen2022edgevit}          &   \Checkmark                &  \Checkmark                 &   \Checkmark               &  \XSolidBrush               &    \Checkmark        \\
% PVT~\cite{wang2021pyramid}          
% & \textcolor{green}{\Checkmark}                 
% & \textcolor{green}{\XSolidBrush}                 
% & \textcolor{red}{\XSolidBrush}                
% & \textcolor{green}{\Checkmark}                 
% & \textcolor{red}{\XSolidBrush}          
% \\
DeiT~\cite{touvron2021training}
& \textcolor{green}{\Checkmark}  
& \textcolor{red}{\XSolidBrush}                  
& \textcolor{red}{\XSolidBrush}                
& \textcolor{green}{\Checkmark}                  
& \textcolor{red}{\XSolidBrush}           
\\
% MobileFormer~\cite{chen2022mobile}          
% & \textcolor{green}{\Checkmark}  
% & \textcolor{green}{\Checkmark}                  
% & \textcolor{green}{\Checkmark}                
% & \textcolor{red}{\XSolidBrush}                  
% & \textcolor{green}{\Checkmark}           
% \\
PoolFormer~\cite{yu2022metaformer}          
& \textcolor{green}{\Checkmark}                  
& \textcolor{green}{\Checkmark}                 
& \textcolor{red}{\XSolidBrush}                
& \textcolor{green}{\Checkmark}                 
& \textcolor{red}{\XSolidBrush}          
\\
RepViT~\cite{wang2024repvit}          
& \textcolor{red}{\XSolidBrush}                  
& \textcolor{green}{\Checkmark}                 
& \textcolor{green}{\Checkmark}                
& \textcolor{red}{\XSolidBrush}                 
& \textcolor{green}{\Checkmark}          
\\
SHViT~\cite{yun2024shvit}          
& \textcolor{green}{\Checkmark}                  
& \textcolor{green}{\Checkmark}                 
& \textcolor{green}{\Checkmark}                
& \textcolor{red}{\XSolidBrush}                 
& \textcolor{green}{\Checkmark}          
\\
\rowcolor{blue!10} \textbf{S2AFormer (ours)}          
& \textcolor{green}{\Checkmark}                  
& \textcolor{green}{\Checkmark}                 
& \textcolor{green}{\Checkmark}                
& \textcolor{green}{\Checkmark}                
& \textcolor{green}{\Checkmark}         
\\
\bottomrule
\end{tabular}
\label{properties}
\vspace{-2em}
\end{center}
\end{table*}

\section{Related Work}
\label {sect:relatedwork}

\subsection{Efficient ViTs}

% The Transformer paradigm has gained widespread attention across the entire field of artificial intelligence. 

Originally developed for long-sequence learning in NLP tasks~\cite{vaswani2017attention}, Transformers were later adapted for image classification% by Dosovitskiy et al.
~\cite{dosovitskiy2020image} and for object detection~\cite{carion2020end}, both achieving performance on par with CNNs due to advanced training techniques and larger datasets. DeiT~\cite{touvron2021training} further improved the training process by incorporating distillation, eliminating the need for extensive pretraining~\cite{yuan2021tokens}. Since then, various modifications of vision Transformers (ViT) and hybrid architectures have emerged, introducing image-specific inductive biases to ViTs, enhancing performance across a range of vision tasks~\cite{xu2024haformer,wang2021pyramid}. While ViTs have outperformed CNNs in several vision tasks due to their token mixer's strong global context capabilities, the complexity of pairwise token interactions and the heavy computational demands of matrix operations present challenges for deploying ViTs in resource-constrained environments and real-time applications.

As a result, researchers have focused on developing various optimization methods to make ViTs more lightweight and better suited for mobile devices. These methods include token pruning or token merging~\cite{kim2024token, tang2023dynamic, rao2021dynamicvit}, the introduction of novel architectures or modules~\cite{shaker2023swiftformer, li2022efficientformer, maaz2022edgenext}, re-evaluating self-attention and sparse-attention mechanisms~\cite{fan2025breaking, ali2021xcit, ma2024efficient}, and applying search algorithms commonly used in CNNs to identify more compact and faster ViT models~\cite{zhou2022training}. 
Notable examples include TPS~\cite{wei2023joint}, which reduces the number of tokens by truncating based on a threshold. This method shortens the sequence length processed by self-attention %by evaluating the relevance score between each token and the class token. However, this approach is most 
and has proved its effectiveness in scenarios where the target objects in the image are relatively simple and singular.
EdgeViT~\cite{chen2022edgevit} improved efficiency by employing a global sparse attention module that focuses on a few selected tokens. In contrast, \cite{wang2021pyramid} achieves a better efficiency-accuracy balance by down-sampling the key and value vectors. EdgeNeXt~\cite{maaz2022edgenext} introduced a transposed self-attention mechanism that computes attention maps along the channel dimension rather than the spatial dimension. This approach, combined with token mixing, results in linear complexity relative to the number of tokens. Reformer~\cite{kitaev2020reformer} used locality-sensitive hashing to group tokens, replacing dot-product attention and reducing complexity from $O(n^2)$ to $O(n log n)$. LinFormer~\cite{wang2020linformer} employed a low-rank matrix factorization technique to approximate the self-attention matrix, reducing complexity from $O(n^2)$ to $O(n)$. Likewise, RAVLT~\cite{fan2025breaking} introduced a rank-augmented linear attention framework that effectively balances the performance of traditional softmax attention with the efficiency of linear attention methods. SwiftFormer~\cite{shaker2023swiftformer} introduces a more efficient additive attention mechanism, replacing quadratic matrix multiplications with linear element-wise operations. 

Although existing methods can alleviate the quadratic computational complexity of self-attention, they often do so at the cost of reduced accuracy. Moreover, approaches that rely exclusively on global self-attention remain insufficient for capturing fine-grained local features, owing to the intrinsic limitations of self-attention in modeling detailed local patterns.

% LightViT~\cite{huang2022lightvit} implements a global aggregation strategy within both token and channel mixers, optimizing the balance between performance and efficiency. 

\begin{figure*}[t]
	\centerline{\includegraphics[width=18cm]{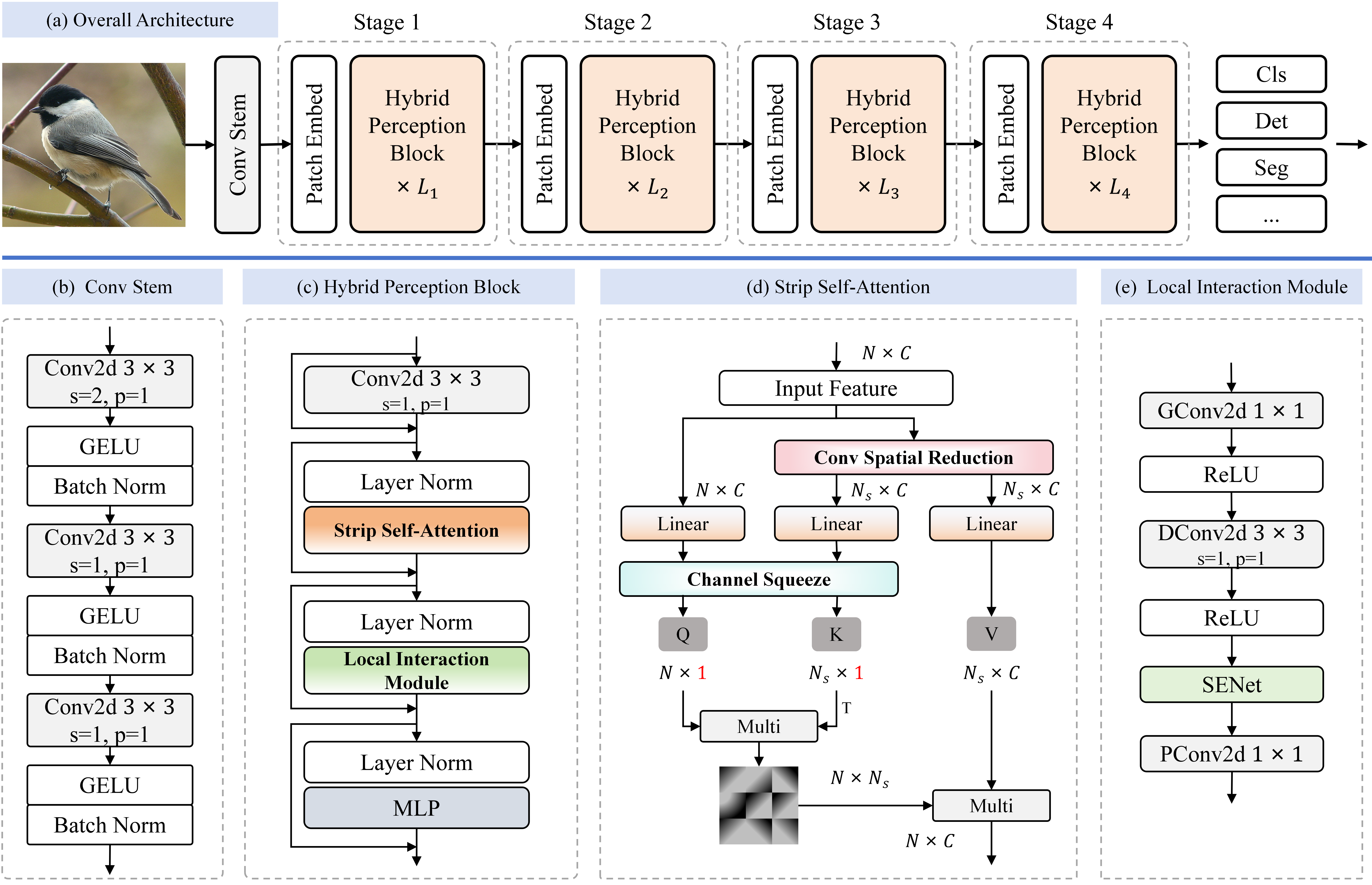}}
	\caption{Overview of our proposed S2AFormer. Similar to~\cite{liu2021swin,wang2021pyramid}, we employ a hierarchical architecture with four stages, each containing $L_i$ Hybrid Perception Blocks. Table~\ref{network_config} provides the detailed network configurations of S2AFormer variants.}
	\label{model}
\end{figure*}

\subsection{Hybrid Variants}

Recent research has introduced a variety of hybrid architectures that integrate CNNs and ViTs within a single framework. By leveraging ViTs' self-attention for capturing long-range dependencies and CNNs' local kernels to preserve fine-grained details, these models have achieved enhanced performance across various vision tasks. MobileViT~\cite{mehta2021mobilevit} employed a hybrid architecture by combining lightweight MobileNet blocks with multi-head self-attention (MHSA) blocks. The MobileFormer~\cite{chen2022mobile} architecture merged MobileNetv3~\cite{koonce2021mobilenetv3} with ViT~\cite{dosovitskiy2020image} to achieve state-of-the-art performance within a 6M-parameter budget, making it well-suited for resource-constrained environments. 
While these methods achieve fast inference speeds and demonstrate strong suitability for edge deployment, their compromised accuracy presents a critical limitation for performance-sensitive applications.

The CMT~\cite{guo2022cmt} architecture introduced a convolutional stem, placing a convolutional layer before each Transformer block, resulting in an alternating sequence of convolutional and Transformer layers. CvT~\cite{wu2021cvt} replaced ViTs' linear embeddings with convolutional token embeddings and uses a convolutional transformer layer to fully leverage these embeddings, ultimately improving performance. However, %they sacrifice inference speed in pursuit of higher accuracy.
their improved accuracy comes at the cost of reduced inference speed. 

Since 2024, %we have observed that, 
to avoid the computational burden and inference speed bottleneck caused by self-attention, efficient methods have been exploring the use of convolution-based techniques as alternatives in the token mixer component. For example, RepViT~\cite{wang2024repvit} built upon MobileNet-V3~\cite{koonce2021mobilenetv3} by streamlining the block structure and removing the residual connection used during training to achieve faster inference speeds. RepNeXt~\cite{zhao2024repnext} utilized multi-scale feature representations while combining both serial and parallel structural reparameterization (SRP) methods to expand the network's depth and width, resulting in a unique convolutional attention mechanism. CAS-ViT~\cite{zhang2024cas} adopted a serial structure with CBAM~\cite{woo2018cbam}, where spatial and channel-wise weights for the query ($Q$) and key ($K$) are calculated. These weights are then summed and used to perform matrix calculations with value ($V$), resulting in an additive attention mechanism. This is a promising approach, given that the computational complexity of convolution is $O(n)$. 
However, purely CNN-based architectures that rely on large-kernel convolutions still fall short in performance compared to self-attention mechanisms, despite offering notable improvements in speed and efficiency.

These observations collectively underscore the essential qualities of an efficient backbone: the ability to capture both local and global features, achieve high performance, and maintain fast inference speed. Making the trade-off between efficiency and precision is a continuing focus of research. As presented in Table~\ref{properties}, the identified properties from successful backbone designs offer important guidance for constructing models that balance accuracy, efficiency, and scalability.

\begin{table}[!t]
\caption{Network configurations of S2AFormer variants.}
\begin{center}
\begin{tabular}{c|cccc}
\toprule
Model & Blocks $L$ & Channels $C$  & \#Para. (M) & GMACs \\  \midrule  
mini &[2,2,2,2]&[32, 64, 128, 256]&5.02 & 0.43\\
T &[2,2,6,2]&[48, 64, 128, 256]&5.80 &0.66\\
XS &[2,2,10,2]&[48, 64, 128, 256]&6.54 &0.79\\
S &[2,4,24,4]&[48, 64, 128, 256]&10.69 &1.38\\
M &[2,4,20,2]&[96, 128, 256, 512]&24.87 & 4.12\\
L &[4,4,20,4]&[96, 192, 384, 768]&76.58 & 12.53\\
\bottomrule
\end{tabular}
\label{network_config}
\vspace{-2em}
\end{center}
\end{table}

\begin{figure*}[htbp]
	\centerline{\includegraphics[width=18cm]{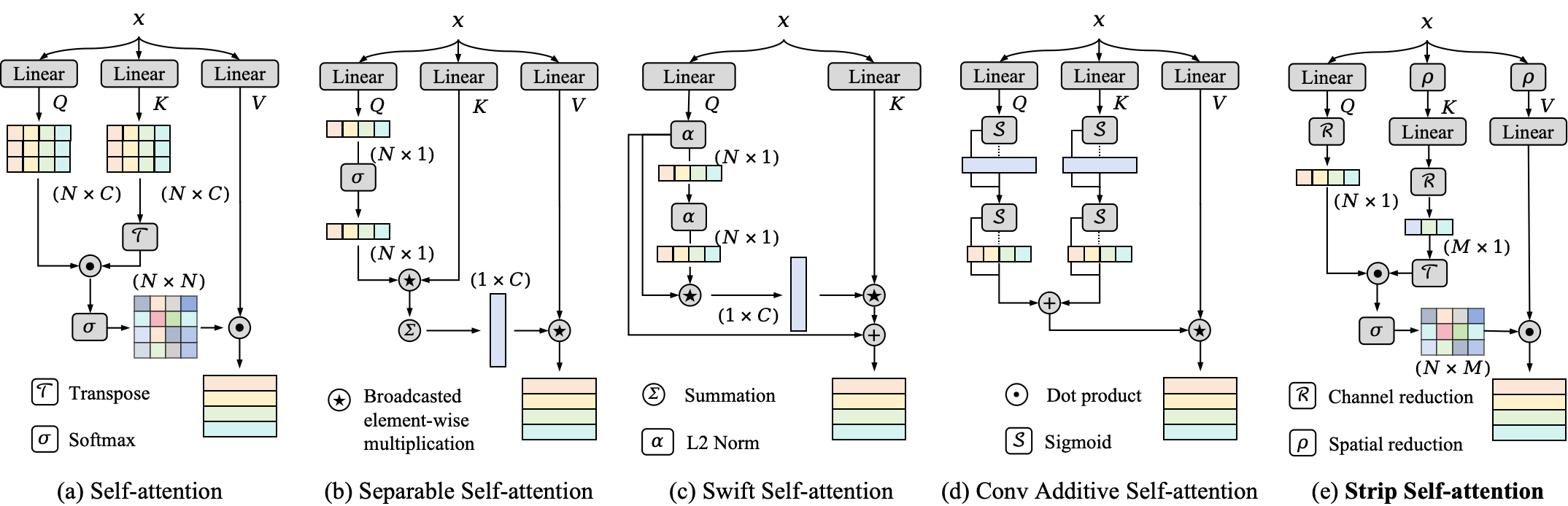}}
	\caption{Comparison of different self-attention mechanisms: (a) Vanilla self-attention in ViTs~\cite{dosovitskiy2020image}, which computes global attention using standard dot-product operations. (b) Separable self-attention in MobileViT-v2~\cite{mehta2022separable}, which applies element-wise operations on query ($Q$) and key ($K$) to form a context vector. (c) Swift self-attention in SwiftFormer~\cite{shaker2023swiftformer}, where $Q$ is weighted and pooled into global queries, broadcast, and multiplied element-wise with $K$ to generate global context. (d) Convolutional additive self-attention in CAS-ViT~\cite{zhang2024cas}, which replaces the global dot-product with a cascaded design applying spatial attention followed by channel attention. (e) Our proposed strip self-attention jointly compresses spatial and channel dimensions to effectively eliminate redundant information, achieving a lightweight design while preserving dense global dependencies.}
	\label{attention}
\end{figure*}

% \begin{figure*}[t]
%     \centering

%     \subfloat[Input image\label{fig:coco4}]{
%         \includegraphics[width=0.2\textwidth]{modelpics/self-attention.png}
%     }\hfill
%     \subfloat[FeedFormer~\cite{shim2023feedformer}\label{fig:feedformer_coco4}]{
%         \includegraphics[width=0.2\textwidth]{"modelpics/separable self-attention.png"}
%     }\hfill
%     \subfloat[VWFormer~\cite{yan2024multi}\label{fig:vwformer_coco4}]{
%         \includegraphics[width=0.2\textwidth]{"modelpics/swift self-attention.png"}
%     }\hfill
%     \subfloat[CCASeg~\cite{yoo2025ccaseg}\label{fig:ccaseg_coco4}]{
%         \includegraphics[width=0.2\textwidth]{"modelpics/strip self-attention.png"}
%     }\hfill
%     % \subfloat{\textbf{WaveSeg (ours)}\label{fig:waveseg_coco4}}{
%     %     \includegraphics[width=0.15\textwidth]{"modelpics/strip self-attention.png"}
%     % }

%     \caption{Qualitative comparisons on the COCO-Stuff 164K~\cite{caesar2018coco}. The yellow boxes highlight regions where our method achieves more precise segmentation.}
%     \label{fig:cocostuff164k} % ← 去掉空格
% \end{figure*}

\section{Methodology}
\label {sect:method}

\subsection{Overall Architecture}

Building on these insights, we propose S2AFormer, an efficient Transformer-based backbone featuring a hybrid architecture that effectively preserves both local and global receptive fields while significantly reducing computational overhead. 
% We aim to design a hybrid architecture that combines the strengths of CNNs and transformers. 
The overall structure of S2AFormer, as illustrated in Fig.~\ref{model}, adopts the four-stage framework commonly used in traditional CNNs~\cite{he2016deep} and hierarchical ViTs~\cite{liu2021swin}, enabling the generation of multi-scale feature maps suitable for dense prediction tasks such as object detection and semantic segmentation. 
%An overview of S2AFormer is illustrated in Fig.~\ref{model}. 

Our approach begins with a Stem Block consisting of three $3 \times 3$ convolutional layers, with the first layer using a stride of 2~\cite{li2023rethinking}. 
This is followed by four stages, each responsible for generating feature maps at different scales. These stages share a consistent architecture, featuring a patch embedding and multiple $L_i$ Hybrid Perception Blocks (HPBs). 
The HPBs, equipped with two core modules, SSA and LIM, effectively capture both local and global dependencies without the heavy computation overhead. % with a detailed explanation provided in Section~\ref{hpb}.

The resulting feature maps, $F_1$, $F_2$, $F_3$, and $F_4$, have strides of 4, 8, 16, and 32 pixels relative to the input image. By leveraging this feature pyramid, % {$F_1$, $F_2$, $F_3$, $F_4$}, 
our approach integrates seamlessly with various downstream tasks, including image classification, object detection, instance segmentation, and semantic segmentation.

\begin{algorithm}[t]
\caption{Pseudocode of the Strip Self-Attention Module}
\label{algorithm_ssa}
\begin{algorithmic}
\Procedure{SSA} {$x, H, W$}
    \State $Q \gets \text{Channel Squeeze} (\text{Linear Projection}(x))$ 
    %\Comment{\textcolor{gray!80}{\textit{Squeeze $Q$'s channel}}}
    \If{$\text{sr\_ratio} > 1$}
        \State $\widetilde{x} \gets \text{Spatial Reduction} (x)$
        \State $K \gets \text{Channel Squeeze} (\text{Linear Projection}(\widetilde{x}))$ 
    %\Comment{\textcolor{gray!80}{\textit{Squeeze $K$'s channel}}}
        \State $V \gets \text{Linear Projection} (\widetilde{x})$
    \Else
        \State $K \gets \text{Channel Squeeze} (\text{Linear Projection}(x))$ 
    %\Comment{\textcolor{gray!80}{\textit{Squeeze $K$'s channel}}}
        \State $V \gets \text{Linear Projection} (x)$
    \EndIf
    \State $attn \gets \text{Softmax} ((Q \cdot K^T) / \text{Scale})$
    \State $y \gets \text{Dropout} (\text{Linear Projection}(attn \cdot V))$
    \State \Return $y$
\EndProcedure
\end{algorithmic}
\end{algorithm}

\subsection{Hybrid Perception Block (HPB)}
\label{hpb}

The Hybrid Perception Block (HPB) framework, shown in Fig.~\ref{model} (c), consists of two main components: the Strip Self-Attention (SSA) module in Fig.~\ref{model} (d) and the Local Interaction Module (LIM) in Fig.~\ref{model} (e). 
The overall operation is defined as follows:
\begin{equation}
    f_{Conv} = DWConv\left( x \right) + x,
\end{equation}
where $x \in {\mathbb{R} ^{H \times W \times C}}$, $H \times W$ represents the resolution of the input at the current stage, while $C$ denotes the feature dimension, and $DWConv$ means the depth-wise convolution. 
The SSA module can be formulated as:
\begin{equation}
    f_{ssa} = SSA\left( {LN\left( f_{Conv} \right)} \right) + f_{Conv},
\end{equation}
where $SSA$ denotes Strip Self-Attention operation, and $LN$ represents the Layer Norm. 

To compensate for the limited local perceptual capacity of self-attention, we integrate a specialized Local Interaction Module, LIM, into the HPB, seamlessly combining global and local receptive fields to improve overall performance. 
The LIM can be formulated as
\begin{equation}
    f_{lim} = LIM\left( {LN\left( f_{ssa} \right)} \right) + f_{ssa}.
\end{equation}

Finally, the MLP is conducted, as:
\begin{equation}
    f_{mlp} = MLP\left( {LN\left( f_{lim} \right)} \right) + f_{lim}.
\end{equation}

Next, we elaborate on the details of these components in the following sections.

\begin{table*}[t]
\caption{Complexity comparison between Multi-Head Self-Attention (MHSA) and Strip Self-Attention (SSA).}
\centering
\renewcommand{\arraystretch}{1.5}
\begin{tabular}{l|c|c|c}
\toprule
\multirow{2}{*}{Stage} & \multirow{2}{*}{Operation} & \multicolumn{2}{c}{Computation Complexity}\\ \cline{3-4}
&& MHSA & SSA\\ \midrule
\multirow{3}{*}{Linear Projection}& $Q=xW_Q \in \mathbb{R}^{N \times d_q}$ & \multirow{3}{*}{$\mathcal{O}(N d (d_q + d_k + d_v)) $} & \multirow{3}{*}{$\mathcal{O}(N d (h + \frac{h}{k^2} + \frac{d_v}{k^2}))$}\\
& $K=xW_K \in \mathbb{R}^{N \times d_k}$ & \\
& $V=xW_V \in \mathbb{R}^{N \times d_v}$  & \\ \midrule
Attention Scores & $Q K^T \in \mathbb{R}^{ N \times N}$ & $\mathcal{O}(N^2d_q)$ & $\mathcal{O}(\frac{N^2}{k^2} \cdot h)$ \\ \midrule
Weighted Sum & $Attn \cdot V \in \mathbb{R}^{N\times d_v}$ & $\mathcal{O}(N^2d_v)$ & $\mathcal{O}(\frac{N^2}{k^2}d_v)$\\ \midrule \midrule
\multirow{3}{*}{Total Complexity} &\multirow{3}{*}{ $h \ll d_q = d_k = d_v = d$} & \multirow{3}{*}{$\mathcal{O}(3N d^2 + 2dN^2)$} & $\mathcal{O}(\frac{1}{k^2}Nd^2 + \frac{d+h}{k^2}N^2+(1+\frac{1}{k^2})Ndh)$\\ 
&&& \multicolumn{1}{l}{$\ll (1+\frac{2}{k^2})Nd^2 + \frac{2}{k^2}dN^2$} \\
&&& \multicolumn{1}{l}{$<3Nd^2 + 2dN^2$} \\

\bottomrule

\end{tabular}
\label{complexity}
\end{table*}

% trim=90 60 40 60
\begin{figure}[t]
    \centering
    % 第一行
    \begin{minipage}[b]{0.15\textwidth}
        \centering
        \includegraphics[width=\linewidth]{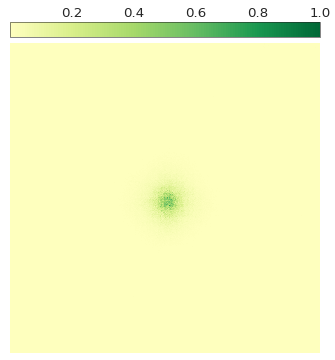}
        \small (a) ResNet~\cite{krizhevsky2012imagenet}
    \end{minipage}
    \begin{minipage}[b]{0.15\textwidth}
        \centering
        \includegraphics[width=\linewidth]{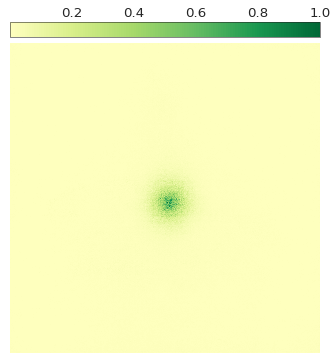}
        \small (b) CAS-ViT~\cite{zhang2024cas}
    \end{minipage}
    \begin{minipage}[b]{0.15\textwidth}
        \centering
        \includegraphics[width=\linewidth]{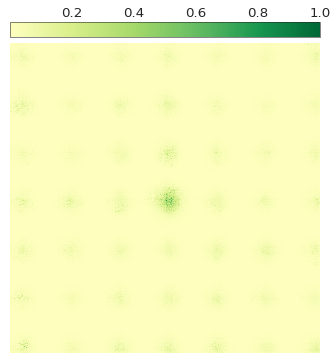}
        \small (c) EMO~\cite{zhang2023rethinking}
    \end{minipage}
    \\
    \vspace{0.5em}
    \begin{minipage}[b]{0.15\textwidth}
        \centering
        \includegraphics[width=\linewidth]{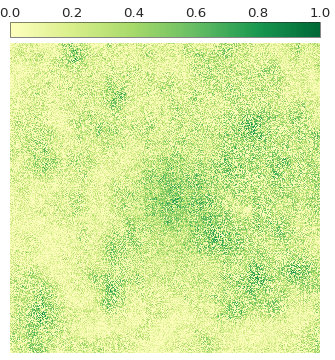}
        \small (d) Vanilla ViT~\cite{dosovitskiy2020image}
    \end{minipage}
    \begin{minipage}[b]{0.15\textwidth}
        \centering
        \includegraphics[width=\linewidth]{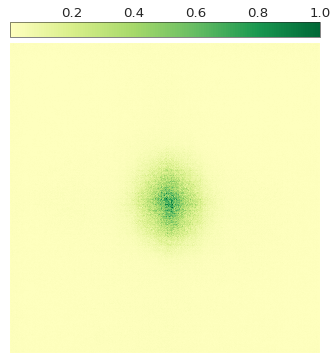}
        \small (e) SwiftFormer~\cite{shaker2023swiftformer}
    \end{minipage}
    \begin{minipage}[b]{0.15\textwidth}
        \centering
        \includegraphics[width=\linewidth]{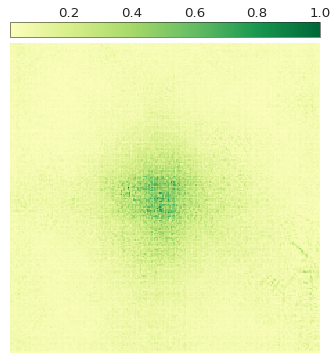}
        \small (f) \textbf{Ours}
    \end{minipage}
    \caption{Visualization of effective respective fields (ERFs) across different models. Convolution-based models (a), (b), and (c) exhibit highly localized receptive fields, while Vanilla ViT (d) distributes attention broadly across all spatial positions. Model (e) demonstrates more limited receptive fields compared to ours. In contrast, our method (f) achieves a balanced pattern—effectively capturing key local regions and progressively expanding outward in a strip-like manner.}
    \label{erf}
    \vspace{-2mm}
\end{figure}

\begin{table*}[t]
\caption{Comparison of image classification performance on the ImageNet-1K dataset~\cite{deng2009imagenet}. \textit{Mixer} refers to the token mixer operation, \textit{Res.} indicates the input resolution of training, and \textit{\#Para.} corresponds to the total number of parameters.}
\centering
\subfloat{
\scalebox{0.8}{
\begin{tabular}{l|lcccc}

\toprule
Model  & Mixer & Res. & \#Para.(M)$\downarrow$& GMACs$\downarrow$ & Top-1(\%)$\uparrow$\\
\toprule
\midrule
MobileFormer-52M~\cite{chen2022mobile} & Conv+Attn & $224^2$ & 3.50 & 0.052 & 68.7 \\
MobileViT-XXS~\cite{mehta2021mobilevit} & Conv+Attn & $256^2$ & 1.30 & 0.364 & 69.0 \\
% MobileNet-v1~\cite{howard2017mobilenets}   & Conv  & $224^2$ & 4.2  & 0.575  & 70.6 \\
MobileViT-v2-0.5~\cite{mehta2022separable} & Attn & $256^2$ & 1.40 & 0.500 & 70.2 \\
EfficientViT-M2~\cite{liu2023efficientvit}  & Conv+Attn & $224^2$ & 4.20 & 0.201 & 70.8 \\
EdgeNeXt-XXS~\cite{maaz2022edgenext} & Conv+Attn & $256^2$ & 1.30 & 0.261 & 71.2 \\
EMO-1M~\cite{zhang2023rethinking} & Conv & $224^2$ & 1.30 & 0.260 & 71.5 \\
FasterNet-T0~\cite{chen2023run} & Conv+Attn & $224^2$ & 3.90 & 0.340 & 71.9 \\
MobileNet-v3~\cite{howard2017mobilenets}   & Conv  & $256^2$ & 1.40  & 0.481  & 72.3 \\
EMOv2-1M~\cite{zhang2025emov2} & Conv & $224^2$ & 1.40 & 0.285 & 72.3 \\
SHViT-S1~\cite{yun2024shvit}  & Conv+Attn & $224^2$ & 6.30 & 0.240  & 72.8 \\
Vim-Ti~\cite{zhu2024vision}  & Conv+SSM & $224^2$ & 7.00 & 1.500  & 73.1 \\
MobileMamba-T2~\cite{he2025mobilemamba} & Conv+SSM & $192^2$ & 8.80 & 0.255 & 73.6 \\
iFormer-T~\cite{zheng2025iformer} & Conv+Attn & $224^2$ & 2.90 & 0.530 & 74.1 \\
LSNet-T~\cite{wang2025lsnet} & Conv+Attn & $224^2$ & 11.40 & 0.300 & 74.9 \\
VRWKV-T~\cite{duan2024vrwkv} & Conv+Attn & $224^2$ & 6.20 & 1.200 & 75.1 \\
\rowcolor{blue!10} \textbf{S2AFormer-mini (Ours)} & \textbf{Conv+Attn} & \textbf{$224^2$} & \textbf{5.02 }& \textbf{0.430}  & \textbf{75.1} \\
\rowcolor{blue!10} \textbf{S2AFormer-mini (Ours)} & \textbf{Conv+Attn} & \textbf{$256^2$} & \textbf{5.02}& \textbf{0.560}  & \textbf{75.7} \\
\midrule
EdgeViT-XXS~\cite{chen2022edgevit} & Conv+Attn & $224^2$ & 4.10 & 0.600 & 74.4 \\
DeiT-T~\cite{touvron2021training}  & Attn  & $224^2$ & 5.90 &  1.200 & 74.5 \\
MobileViT-XS~\cite{mehta2021mobilevit} & Conv+Attn & $224^2$ & 2.30 & 0.706 & 74.8 \\
SHViT-S2~\cite{yun2024shvit}  & Conv+Attn & $224^2$ & 11.40 & 0.370  & 75.2 \\
InceptionNeXt-A~\cite{yu2024inceptionnext} & Conv & $224^2$ &4.20 & 0.510 & 75.3 \\
FastViT-T8~\cite{vasu2023fastvit}  & Conv+Attn & $256^2$ & 3.60 & 0.700 & 75.6 \\
% EfficientFormerV2-S0~\cite{li2023rethinking}  & Conv+Attn & $224^2$ & 3.5 & 0.4  & 75.7 \\
EMOv2-2M &  Conv & $224^2$ & 2.30 & 0.487 & 75.8 \\
MobileMamba-T4~\cite{he2025mobilemamba} & Conv+SSM & $192^2$ & 14.20 & 0.413 & 76.1 \\
% LocalVim-T~\cite{huang2024localmamba}  & Conv+SSM & $224^2$ & 8.00 & 1.500  & 76.2 \\
EfficientVMamba-T~\cite{pei2024efficientvmamba}  & Conv+SSM & $224^2$ & 6.00 & 0.800 & 76.5 \\
EfficientViT-M5~\cite{liu2023efficientvit}  & Conv+Attn & $224^2$ & 12.40 & 0.522 & 77.1 \\
PoolFormer-S12~\cite{yu2022metaformer}   & Pool  & $224^2$ & 12.00 & 1.900  & 77.2 \\
SHViT-S3~\cite{yun2024shvit}  & Conv+Attn & $224^2$ & 14.20 & 0.600  & 77.4 \\
LSNet-S~\cite{wang2025lsnet} & Conv+Attn & $224^2$ & 16.10 & 0.500 & 77.8 \\
FAT-B0~\cite{fan2023lightweight} & Conv+Attn & $224^2$ & 4.50 & 0.700 & 77.6 \\
% \rowcolor{gray!10} \textbf{S2AFormer-nano (Ours)} & \textbf{Conv+Attn} & \textbf{$224^2$} & \textbf{3.2}& \textbf{0.6}  & \textbf{76.7} \\
\rowcolor{blue!10} \textbf{S2AFormer-T (Ours)} & \textbf{Conv+Attn} & \textbf{$224^2$} & \textbf{5.80 }& \textbf{0.655}  & \textbf{77.7} \\
\rowcolor{blue!10} \textbf{S2AFormer-T (Ours)} & \textbf{Conv+Attn} & \textbf{$256^2$} & \textbf{5.80} & \textbf{0.855}  & \textbf{78.3} \\
\midrule

MobileViG-S~\cite{munir2023mobilevig} & Conv+GNN & $224^2$ & 7.20 & 1.000  & 78.2 \\
EdgeViT-XS~\cite{chen2022edgevit}  & Conv+Attn & $224^2$ & 6.70 & 1.100 & 77.5 \\
MobileFormer-294M~\cite{chen2022mobile} & Conv+Attn & $224^2$ & 11.40 & 0.294 & 77.9 \\
SwiftFormer-S~\cite{shaker2023swiftformer}  & Conv+Attn & $224^2$ & 6.10 & 1.000 & 78.5 \\

% PlainMamba-L1~\cite{yang2024plainmamba}  & Conv+SSM & $224^2$ & 7 & 3.0  & 77.9 \\
MobileViT-v2-1.0~\cite{mehta2022separable} & Attn & $256^2$ & 4.90 & 1.800 & 78.1 \\
EdgeNeXt-S~\cite{maaz2022edgenext}  & Conv+Attn & $224^2$ & 5.60 & 0.965 & 78.8 \\
MobileMamba-S6~\cite{he2025mobilemamba} & Conv+SSM & $224^2$ & 15.00 & 0.652 & 78.0 \\
%\color{blue}iFormer-S~\cite{zheng2025iformer} & \color{blue}Conv+Attn & \color{blue}$224^2$ & \color{blue}6.50 & \color{blue}1.090 & \color{blue}78.8 \\
RepViT-M0.9~\cite{wang2024repvit}  & Conv & $224^2$ & 5.10 & 0.800  & 78.7 \\
FastViT-T12~\cite{vasu2023fastvit}  & Conv+Attn & $256^2$ & 6.80 & 1.400 & 79.1 \\
EfficientFormer-L1~\cite{li2022efficientformer}  & Conv+Attn & $224^2$ & 12.30 & 1.300 & 79.2\\
EfficientMod-xs~\cite{ma2024efficient}  & Conv & $224^2$ & 6.60 & 0.800  & 78.3 \\
EfficientVMamba-S~\cite{pei2024efficientvmamba}  & Conv+SSM & $224^2$ & 11.00 & 1.300  & 78.7 \\
MobileViT-S~\cite{mehta2021mobilevit}  & Conv+Attn & $224^2$ & 5.60 & 2.010  & 78.4 \\
StarNet-S4~\cite{ma2024rewrite}  & Conv & $224^2$ & 7.50 & 1.075  & 78.4 \\
\rowcolor{blue!10} \textbf{S2AFormer-XS (Ours)}  & \textbf{Conv+Attn} & \textbf{$224^2$} & \textbf{6.54} & \textbf{0.786}   & \textbf{78.9} \\
\rowcolor{blue!10} \textbf{S2AFormer-XS (Ours)}  & \textbf{Conv+Attn} & \textbf{$256^2$} & \textbf{6.54} & \textbf{1.030}   & \textbf{79.3} \\

\bottomrule
\end{tabular}
}
}
% \midrule
\subfloat{
\scalebox{0.8}{
\begin{tabular}{l|lcccc}

\toprule
Model  & Mixer & Res. & \#Para.(M)$\downarrow$& GMACs$\downarrow$ & Top-1(\%)$\uparrow$\\
\toprule
\midrule
EdgeNeXt-S~\cite{maaz2022edgenext} & Conv+Attn & $256^2$ & 5.60 & 1.300  & 79.4 \\
SHViT-S4~\cite{yun2024shvit}  & Conv+Attn & $256^2$ & 16.50 & 0.990  & 79.4 \\
MobileMamba-B1~\cite{he2025mobilemamba} & Conv+SSM & $256^2$ & 17.10 & 1.080 & 79.9 \\
PoolFormer-S24~\cite{yu2022metaformer} & Pool  & $224^2$ & 21.00 & 3.500 & 80.3 \\
% EfficientFormerV2-S2  & Conv+Attn & $224^2$ & 12.6 & 1.25  & 81.6\\
FAT-B1~\cite{fan2023lightweight} & Conv+Attn & $224^2$ & 7.80 & 1.200 & 80.1 \\
VRWKV-S~\cite{duan2024vrwkv} & Conv+Attn & $224^2$ & 23.80 & 4.600 & 80.1 \\
Vim-S~\cite{zhu2024vision}   & Conv+SSM & $224^2$ & 26.00 & 5.100  & 80.3 \\
LSNet-B~\cite{wang2025lsnet} & Conv+Attn & $224^2$ & 23.20 & 1.300 & 80.3 \\
MobileViG-M~\cite{munir2023mobilevig} & Conv+GNN & $224^2$ & 14.00 & 1.500  & 80.6 \\
SwiftFormer-L1~\cite{shaker2023swiftformer} & Conv+Attn & $224^2$ & 12.10 & 1.620 & 80.9 \\
EfficientViT-M5~\cite{liu2023efficientvit}  & Conv+Attn & $512^2$ & 12.40 & 2.670 & 80.8 \\
FastViT-SA12~\cite{vasu2023fastvit}  & Conv+Attn & $256^2$ & 10.90 & 1.900 & 80.6 \\
RepViT-M1.1~\cite{wang2024repvit} & Conv & $224^2$ & 8.20 & 1.300 & 80.7 \\
EdgeViT-S~\cite{chen2022edgevit}  & Conv+Attn & $224^2$ & 11.10 & 1.900  & 81.0 \\
EfficientMod-s~\cite{ma2024efficient} & Conv & $224^2$ & 12.90 & 1.400  & 81.0 \\
\rowcolor{blue!10} \textbf{S2AFormer-S (Ours)}  & \textbf{Conv+Attn} & \textbf{$224^2$} & \textbf{10.69}  & \textbf{1.380}    & \textbf{80.8}  \\
\rowcolor{blue!10} \textbf{S2AFormer-S (Ours)}  & \textbf{Conv+Attn} & \textbf{$256^2$} & \textbf{10.69} & \textbf{1.800}  & \textbf{81.3}  \\

\midrule
RMT-T~\cite{fan2024rmt} & Conv & $224^2$ & 14.00 & 2.500  & 82.4 \\
TransNeXt-Micro~\cite{shi2024transnext} & Conv+Attn & $224^2$ & 12.80 & 2.700  & 82.5 \\
EfficientFormer-L3~\cite{li2022efficientformer} & Conv+Attn & $224^2$ & 31.30 & 3.900  & 82.4 \\
% EfficientFormerV2-L  & Conv+Attn & $224^2$ & 26.1 & 2.56  & 83.3\\
MobileViG-Ti~\cite{munir2023mobilevig}  & Conv+GNN & $224^2$ & 26.70 & 2.800 & 82.6 \\
% SwiftFormer-L3  & Conv+Attn & $224^2$ & 28.5 & 4.0 & 83.0 \\
% FastViT-SA24  & Conv+Attn & $256^2$ & 20.6 & 3.8 & 82.6 \\
% FastViT-SA36 & Conv+Attn & $256^2$ & 30.4 & 5.6  & 83.6 \\
RepViT-M1.5~\cite{wang2024repvit} & Conv & $224^2$ & 14.00 & 2.300  & 82.3 \\
% RepNeXt-M4~\cite{zhao2024repnext} & Conv & $224^2$ & 13.3 & 2.3  & 81.2\\
CAS-ViT-M~\cite{zhang2024cas} & Conv & $224^2$ & 12.42 & 1.887  & 81.4 \\
MobileMamba-B2~\cite{he2025mobilemamba} & Conv+SSM & $384^2$ & 17.10 & 2.427 & 81.6 \\
% \color{blue}FAT-B2~\cite{fan2023lightweight} & \color{blue}Conv+Attn & \color{blue}$224^2$ & \color{blue}13.50 & \color{blue}2.000 & \color{blue}81.9 \\
VRWKV-B~\cite{duan2024vrwkv} & Conv+Attn & $224^2$ & 93.70 & 18.200 & 82.0 \\
SHViT-S4~\cite{yun2024shvit}  & Conv+Attn & $512^2$ & 16.50 & 3.970  & 82.0 \\
% VMamba-T~\cite{liu2024vmamba}  & Conv+SSM & $224^2$ & 22 & 5.6  & 82.2 \\
MobileMamba-B4~\cite{he2025mobilemamba} & Conv+SSM & $512^2$ & 17.10 & 4.313 & 82.5 \\
MambaOut-Tiny~\cite{yu2024mambaout} & Conv & $224^2$ & 27.00 & 4.500 & 82.7 \\
EfficientVMamba-B~\cite{pei2024efficientvmamba} & Conv+SSM & $224^2$ & 33.00 & 4.000  & 81.8 \\
% LocalVim-S~\cite{huang2024localmamba}  & Conv+SSM & $224^2$ & 28 & 4.8 & 81.2 \\
InceptionNeXt-T~\cite{yu2024inceptionnext} & Conv & $224^2$ & 28.00 & 4.200 & 82.3 \\
PeLK-T~\cite{chen2024pelk} & Conv & $224^2$ & 29.00 & 5.600 & 82.6 \\
\rowcolor{blue!10} \textbf{S2AFormer-M (Ours)}  & \textbf{Conv+Attn} & \textbf{$224^2$} & \textbf{24.87} & \textbf{4.120}   & \textbf{82.3} \\
\rowcolor{blue!10} \textbf{S2AFormer-M (Ours)}  & \textbf{Conv+Attn} & \textbf{$256^2$} & \textbf{24.87} & \textbf{5.380}   & \textbf{82.7} \\
\midrule

ConvNeXt-B~\cite{liu2022convnet} & Conv & $224^2$ & 89.00 & 15.400  & 83.8 \\
Swin-B~\cite{liu2021swin} & Attn & $224^2$ & 88.00 & 15.400  & 83.5 \\
EfficientFormer-L7~\cite{li2022efficientformer} & Conv+Attn & $224^2$ & 82.10 & 10.200  & 83.3 \\
% FastViT-SA36 & Conv+Attn & $384^2$ & 30.4 & 12.6 & 84.5 \\
PlainMamba-L3~\cite{yang2024plainmamba}  & Conv+SSM & $224^2$ & 50.00 & 14.400 & 82.3 \\
Focal-Base~\cite{yang2021focal} & Attn & $224^2$ & 89.80 & 16.000 & 83.8 \\
% PvTv2-b4 & Attn & $224^2$ & 63.0 & 10.0 & 83.6 \\
% MViTv2-B & Attn & $224^2$ & 52.0 & 10.2 & 84.4 \\
GroupMixFormer-B~\cite{ge2023advancing} & Attn & $224^2$ & 45.80 & 17.600 & 84.7 \\
CAFormer-M36~\cite{yu2024metaformer} & Attn & $224^2$ & 56.00 & 13.200 & 85.2 \\
VRWKV-L~\cite{duan2024vrwkv} & Conv+Attn & $384^2$ & 334.90 & 189.500 & 86.0 \\
RAVLT-L~\cite{fan2025breaking} &Attn & $224^2$ & 95.00 & 16.000 & 85.8 \\
XCiT-M24/16~\cite{ali2021xcit} & Attn & $224^2$ & 84.00 & 16.200 & 84.3 \\
InceptionNeXt-B~\cite{yu2024inceptionnext} & Conv & $224^2$ & 87.00 & 14.900 & 84.0 \\
PoolFormer-M48~\cite{yu2022metaformer} & Conv & $224^2$ & 73.00 & 11.600 & 82.5 \\
%\color{blue}SG-Former-B~\cite{ren2023sgformer} & \color{blue}Attn & \color{blue}$224^2$ & \color{blue}77.90 & \color{blue}15.600 & \color{blue}84.7 \\
Conv2Former-B~\cite{hou2024conv2former} & Conv+Attn & $224^2$ & 90.00 & 15.900 & 84.4 \\
TransNeXt-Base~\cite{shi2024transnext} & Conv+Attn & $224^2$ & 89.70 & 18.400 & 84.8 \\
\rowcolor{blue!10} \textbf{S2AFormer-L (Ours)}  & \textbf{Conv+Attn} & \textbf{$224^2$} & \textbf{76.58}  & \textbf{12.530}    &\textbf{85.6}   \\
\rowcolor{blue!10} \textbf{S2AFormer-L (Ours)}  & \textbf{Conv+Attn} & \textbf{$256^2$} &\textbf{76.58}  & \textbf{16.370}   &\textbf{86.0}  \\

\bottomrule

\end{tabular}
}

}
\label{imagenet1k}
\vspace{-2em}
\end{table*}

\subsubsection{Strip Self-Attention (SSA)}

In the standard self-attention mechanism~\cite{dosovitskiy2020image}, the input $x \in {\mathbb{R} ^{N \times C}}$, where $N=HW$, is linearly projected into the query $Q \in {\mathbb{R} ^{N \times d_q}}$, the key $K\in {\mathbb{R} ^{N \times d_k}}$, and the value $ V\in {\mathbb{R} ^{N \times d_v}}$. Subsequently, the self-attention mechanism is executed as follows: 
\begin{equation}
    Attn\left( {Q,K,V} \right) = Softmax \left( {\frac{{Q{K^T}}}{{\sqrt {{d_k}} }}} \right)V.
\end{equation}

To reduce computational complexity, a Convolution Spatial Reduction (CSR)~\cite{wang2021pyramid} is employed, which uses a $k \times k$ depth-wise convolution with a stride of $k$ to downsample the spatial dimensions of $x$ before the linear projection operation. This results in $\widetilde{x} \in {\mathbb{R} ^{N_s \times C}}$, where $N_s = \frac{N}{{{k^2}}}$. In our work, we propose a Strip Self-Attention (SSA) module as an innovative token mixing mechanism in the HPB to balance global feature extraction and the computational efficiency of self-attention. 

As shown in Fig.~\ref{model} (d), the channel dimensions of the query $Q$ and key $K$ are compressed into a single dimension to further minimize computational overhead. 
Experimental results demonstrate that a strip-shaped configuration for $Q \in {\mathbb{R} ^{N \times 1}}$ and $K \in {\mathbb{R} ^{N_s \times 1}}$ can effectively capture global similarities. 
The pseudocode for the entire SSA operation is presented in Algorithm~\ref{algorithm_ssa}, followed by a detailed complexity analysis. 

Fig.~\ref{attention} compares the proposed Strip Self-Attention with several notable self-attention variants from prior works. 
(a) shows the vanilla self-attention used in ViTs~\cite{dosovitskiy2020image}, which computes global attention via standard dot-product operations. 
(b) depicts the separable self-attention in MobileViT-v2~\cite{mehta2022separable}, which forms a context vector via element-wise interactions between query ($Q$) and key ($K$), then multiplies it with value ($V$) to produce the output. 
(c) presents the swift self-attention in SwiftFormer~\cite{shaker2023swiftformer}, an additive variant that relies solely on $Q$ and $K$: the query matrix is reweighted with learnable parameters, pooled into global queries, broadcast, and multiplied element-wise with $K$ to obtain a global context. 
(d) illustrates the convolutional additive self-attention in CAS-ViT~\cite{zhang2024cas}, which replaces the core token mixer with a sequence of convolutional spatial attention followed by channel attention, inspired by CBAM~\cite{woo2018cbam}. 
Our Strip Self-Attention differs in two key respects. Firstly, the mechanisms in (b) and (c) operate at the attention-map level using element-wise products rather than full matrix attention, limiting their ability to model true global context. Secondly, CAS-ViT’s~\cite{zhang2024cas} purely convolutional path struggles to capture dense token correlations. In contrast, our method jointly reduces redundancy in spatial and channel dimensions, minimizing computational load while preserving global awareness and effectively modeling dense token interactions.

To facilitate clearer understanding, the effective receptive fields (ERFs) of various architectures are distinctly visualized in Fig.~\ref{erf}. Convolution-based methods display highly localized receptive fields, which restrict their ability to capture global context. In contrast, the Vanilla ViT~\cite{dosovitskiy2020image} covers all spatial positions, allowing for comprehensive global information extraction, but at the cost of significantly increased computational complexity. Our method achieves a balance between these two extremes. As illustrated, the receptive field in our approach initially focuses on the most critical regions and then progressively expands in a strip-like manner along the four cardinal directions, effectively emulating human visual perception mechanisms.

\subsubsection{Local Interaction Module (LIM)}
Vision Transformers (ViTs) face two main limitations compared to CNNs. First, absolute positional encoding disrupts translation invariance by assigning unique encodings to each patch. This issue can be mitigated by using relative or no positional encoding. Second, ViTs often overlook local relationships and structural information within patches. To overcome this, we propose a Local Interaction Module (LIM) to effectively capture local features, which is formulated as
\begin{equation}
    {f_{ds}} = DWConv\left( {\sigma \left( {PWConv\left( f \right)} \right)} \right),
\end{equation}
where $DWConv$ refers to depth-wise convolution, while $PWConv$ represents point-wise convolution, together forming the structure of depth-wise separable convolution. The symbol $\sigma$ typically denotes the activation function $ReLU$. 
The final output of LIM can be expressed as:
\begin{equation}
    {f_{loc}} = PWConv\left( {\sigma \left( {SE\left( {{f_{ds}}} \right)} \right)} \right),
\end{equation}
where $SE$ denotes the channel attention module~\cite{hu2018squeeze}.

Due to its depth-wise architectural design, the proposed LIM has an almost negligible computational footprint. It has minimal parameter overhead and minimal impacts on inference throughput and memory efficiency.

%A comparison of the receptive field characteristics and information interaction patterns for different strategies is presented in Fig~\ref{percaption}. Convolutional networks exhibit strong local sensitivity due to their limited kernel size, enabling precise capture of fine-grained details but lacking global context. In contrast, vanilla Transformers model global dependencies through dense token-to-token interactions, achieving a wide receptive field at the expense of computational efficiency and limited ability to resolve local structures. Our approach effectively integrates the complementary advantages of both paradigms by selectively attending to semantically important regions while suppressing irrelevant information, thereby achieving a better trade-off between expressiveness and efficiency in alignment with human visual perception.

\subsection{Complexity Analysis}

To demonstrate the efficiency of our proposed SSA, we compared its computational overhead with standard self-attention~\cite{dosovitskiy2020image}. 
Table~\ref{complexity} presents a detailed computational complexity comparison of the two attention mechanisms.

Specifically, given an input feature $x \in \mathbb{R}^{N \times d}$, %we analyze the computational cost, measured in FLOPs, as follows. 
the total computational cost of Multi-Head Self-Attention (MHSA), measured in FLOPs, is given by
\begin{equation}
    \mathcal{O}(\text{MHSA}) = N d (d_q + d_k + d_v) + N^2 (d_q + d_v).
\end{equation}
%
% The projection step introduces the complexity of 
% \begin{equation}
%     \mathcal{O}(\text{proj}) = N d (d_q + d_k + d_v),
% \end{equation}
% where $d_q$, $d_k$, and $d_v$ represent the dimensions of the query, key, and value, respectively. Calculating the attention weights involves the dot product of $Q$ and $K^T$, which incurs a complexity of 
% \begin{equation}
%     \mathcal{O}(Q \cdot K^T) = N^2 \cdot d_q.
% \end{equation}
% Subsequently, applying the attention weights to $V$ requires 
% \begin{equation}
%     \mathcal{O}(\text{Attn} \cdot V) = N^2 \cdot d_v.
% \end{equation}
% Combining these, the total computational cost of Multi-Head Self-Attention (MHSA) is given by 
%

When assuming $d_q = d_k = d_v = d$, this simplifies to 
\begin{equation}
    \mathcal{O}(\text{MHSA}) = 3N d^2 + 2N^2 d.
\end{equation}

In the case of our SSA, a spatial reduction is introduced, with $N_s = \frac{N}{k^2}$, where $k$ represents the kernel size of the spatial reduction convolution. 

% The projection phase, incorporating the effects of spatial reduction, has a complexity of 
% \begin{equation}
%     \Omega(\text{proj}) = N d \left(h + \frac{h}{k^2} + \frac{d}{k^2}\right),
% \end{equation}
% where $h \ll d$ is the number of attention heads. The computation of attention weights involves 
% \begin{equation}
%     \Omega(Q \cdot K^T) = N \cdot N_s \cdot h = \frac{N^2}{k^2} \cdot h.
% \end{equation}
% Similarly, multiplying the attention weights by $V$ results in a complexity of 
% \begin{equation}
%     \Omega(\text{Attn} \cdot V) = N N_s \cdot d = \frac{N^2 d}{k^2}.
% \end{equation}
Thus, the total computational cost for SSA is expressed as 
\begin{equation}
    \mathcal{O}(\text{SSA}) = \frac{1}{k^2}N d^2 + \frac{h + d}{k^2}N^2 + \left(1 + \frac{1}{k^2}\right) N d h.
\end{equation}
For $h \ll d$, the dominant terms simplify to 
\begin{equation}
    \mathcal{O}(\text{SSA}) \ll \frac{1}{{{k^2}}}N{d^2} + \frac{{2d}}{{{k^2}}}{N^2} + \left( {1 + \frac{1}{{{k^2}}}} \right)N{d^2}.
\end{equation}
Let $\mathcal{O}(\text{A}) = \frac{1}{{{k^2}}}N{d^2} + \frac{{2d}}{{{k^2}}}{N^2} + \left( {1 + \frac{1}{{{k^2}}}} \right)N{d^2}$, the complexity can be further simplified as
\begin{equation}
    \begin{aligned}
        \mathcal{O}(\text{A}) 
        &= \left(1 + \frac{2}{k^2}\right) N d^2 + \frac{2d}{k^2}N^2 \\
        &< 3N d^2 + 2dN^2.
    \end{aligned}
    \label {eq18}
\end{equation}

Eq.~\ref{eq18} shows that the computational cost of SSA is significantly lower than that of MHSA, especially when the reduction factor $k$ is large, satisfying 
\begin{equation}
    \mathcal{O}(\text{SSA}) \ll \mathcal{O}(\text{A}) < \mathcal{O}(\text{MHSA}).
\end{equation}

\begin{table}[!t]
\caption{Comparison of semantic segmentation performance on the ADE20K dataset~\cite{zhou2017scene}.}
\begin{center}
\scalebox{0.9}{
\begin{tabular}{l|ccc}
\toprule
\multicolumn{1}{l|}{\multirow{2}{*}{Model}} & \multicolumn{3}{c}{Semantic FPN~\cite{kirillov2019panoptic}}          \\
\multicolumn{1}{c|}{}                          & \multicolumn{1}{c}{\#Para.(M)$\downarrow$} & \multicolumn{1}{c}{GFLOPs$\downarrow$} & \multicolumn{1}{c}{mIoU(\%)$\uparrow$} \\ \midrule
\midrule

ResNet-18~\cite{he2016deep} & - & - & 32.9 \\
EMO-1M~\cite{zhang2023rethinking} & 5 & 23 & 34.2 \\
PVT-T~\cite{wang2021pyramid} & 17 & 33 &35.7 \\
ResNet-50~\cite{he2016deep} & 29 & 46 & 36.7 \\
\rowcolor{blue!10} \textbf{S2AFormer-mini (Ours)} & \textbf{6}  & \textbf{23}   & \textbf{36.7}  \\ \midrule

PoolFormer-S12~\cite{yu2022metaformer} & 16 & 31 & 37.2 \\
PVTv2-B0~\cite{wang2021pvtv2} &8 &25 &37.2 \\
EMO-2M~\cite{zhang2023rethinking} & 6 & 24 & 37.3 \\
CASViT-XS~\cite{zhang2024cas} & 7 & 24 & 37.1 \\
\rowcolor{blue!10} \textbf{S2AFormer-T (Ours)} & \textbf{7}  & \textbf{25}   & \textbf{38.0}  \\ \midrule
FastViT-SA12~\cite{vasu2023fastvit} & 14 & 29 & 38.0 \\
ResNet-101~\cite{he2016deep} & 48 & 65 & 38.8 \\
EfficientFormer-L1~\cite{li2022efficientformer} & 16 & 28 & 38.9 \\
ResNeXt-101-32x4d~\cite{xie2017aggregated} & 47 & 65 & 39.7 \\
% EdgeViT-XXS~\cite{chen2022edgevit} & 8 & 24 & 39.7 \\
\rowcolor{blue!10} \textbf{S2AFormer-XS (Ours)} & \textbf{8}  & \textbf{26}   & \textbf{39.2}  \\ \midrule
LSNet-T~\cite{wang2025lsnet} & - & - & 40.1 \\
ResNeXt-101-64x4d~\cite{xie2017aggregated} & 86 & 104 & 40.2 \\
EMO-5M~\cite{zhang2023rethinking} & 9 & 26 & 40.4 \\
PVT-S~\cite{wang2021pyramid} & 28 & 45 & 39.8 \\
RepViT-M1.1~\cite{wang2024repvit} & - & - & 40.6 \\
%\color{blue}RepNeXt-M3~\cite{zhao2024repnext} & \color{blue}- & \color{blue}- & \color{blue}40.6 \\
% FastViT-SA24~\cite{vasu2023fastvit} & 24 & 37 & 41.0 \\
PoolFormer-S24~\cite{yu2022metaformer} & 25 & 39 & 40.3 \\
\rowcolor{blue!10} \textbf{S2AFormer-S (Ours)} & \textbf{12}  & \textbf{28}   & \textbf{40.8}  \\ \midrule
SwiftFormer-L1~\cite{shaker2023swiftformer} & 16 & 30 & 41.4 \\
EfficientViM-M4~\cite{lee2025efficientvim} & - & - & 41.3 \\
PoolFormer-S36~\cite{yu2022metaformer} & 35 & 48 & 42.0 \\
EfficientFormerV2-S2~\cite{li2023rethinking} & 16 & 28 & 42.4 \\
iFormer-M~\cite{zheng2025iformer} & - & - & 42.4 \\
PVTv2-B1~\cite{wang2021pvtv2} & 18 & 34 & 42.5 \\
MobileMamba-B4~\cite{he2024mobilemamba} & 20 & - & 42.5 \\
PoolFormer-M48~\cite{yu2022metaformer} & 77 & 82 & 42.7 \\
InceptionNeXt-T~\cite{yu2024inceptionnext} & 28 & 44 & 43.1 \\
% RepNeXt-M4~\cite{zhao2024repnext} & - & - & 43.3 \\
FastViT-SA36~\cite{vasu2023fastvit} & 34 & 44 & 42.9 \\
LSNet-B~\cite{wang2025lsnet} & - & - & 43.0 \\
EfficientMod-S~\cite{ma2024efficient} & 33 & - & 43.5 \\
% EfficientFormer-L3~\cite{li2022efficientformer} & - & - & 43.5\\
%\color{blue}VRWKV-T~\cite{duan2024vrwkv} & \color{blue}- & \color{blue}- & \color{blue}43.3 \\
%RepViT-M1.5~\cite{wang2024repvit} & - & - & 43.6 \\
% CASViT-M~\cite{zhang2024cas} & 15.8 & 31.3 & 43.6 \\

\rowcolor{blue!10} \textbf{S2AFormer-M (Ours)} & \textbf{26}  & \textbf{43}   & \textbf{43.7}  \\   \bottomrule                           
\end{tabular}
\label{segmentation}
\vspace{-3em}
}
\end{center}
\end{table}

\begin{table*}[t]
    \renewcommand\arraystretch{1.0}
    \begin{center}
    \tabcolsep=1.8mm
    \caption{Comparison of object detection and instance segmentation performance (\%) on the COCO val2017 dataset~\cite{lin2014microsoft}. FLOPs are tested on images of size $800 \times 1280$. Our results, highlighted in bold, demonstrate superior performance with comparative computational overhead.}
    \label{table:det}
    % \resizebox{0.9\linewidth}{!}{
        \begin{tabular}{l|cc|cccccc|cccccc}
        \toprule
        
        \multirow{2}{*}{Model} & \multirow{2}{*}{\#Para.(M)$\downarrow$} & \multirow{2}{*}{GFLOPs$\downarrow$} & \multicolumn{6}{c|}{RetinaNet 1$\times$} & \multicolumn{6}{c}{Mask R-CNN 1$\times$} \\
        \cline{4-15}
        &  &  & AP & AP$_{50}$ & AP$_{75}$ & AP$_S$ & AP$_M$ & AP$_L$ & AP$^b$ & AP$^b_{50}$ & AP$^b_{75}$ & AP$^m$ & AP$^m_{50}$ & AP$^m_{75}$ \\
        \midrule
        \midrule
        MobileNetV2~\cite{sandler2018mobilenetv2} & -/-  &-/- &28.3 &46.7 &29.3 &14.8 &30.7 &38.1 &29.6 &48.3 &31.5 &27.2 &45.2 &28.6\\
        MobileNetV3~\cite{koonce2021mobilenetv3} & -/-  &-/- &29.9 &49.3 &30.8 &14.9 &33.3 &41.1 &29.2 &48.6 &30.3 &27.1 &45.5 &28.2 \\
        
        ResNet-18~\cite{he2016deep} & 21/31 & -/- & 31.8 & 49.6 & 33.6 & 16.3 & 34.3 & 43.2 & 34.0 & 54.0 & 36.7 & 31.2 & 51.0 & 32.7 \\
        
        EfficientViT-M4~\cite{liu2023efficientvit}& 9/- & -/- & 32.7 & 52.2 & 34.1 & 17.6 & 35.3 & 46.0 &32.8 &54.4 &34.5 &31.0 &51.2 & 32.2 \\
        \rowcolor{blue!10}\textbf{S2AFormer-mini (Ours)} & \textbf{12/22} & \textbf{159/177} & \textbf{33.4} & \textbf{53.2} & \textbf{34.9} & \textbf{19.9} & \textbf{36.3} & \textbf{44.5} & \textbf{33.4} &  \textbf{55.4} & \textbf{35.2} & \textbf{31.7} & \textbf{52.5} & \textbf{33.3} \\ \midrule
        LSNet-T~\cite{wang2025lsnet} &  -/- &  -/- &  34.2 &  54.6 &  35.2 &  17.8 &  37.1 &  48.5 &  35.0 &  57.0 &  37.3 &  32.7 &  53.8 &  34.3 \\
        PoolFormer-S12~\cite{yu2022metaformer}  & 22/32  & -/- & 36.2 & 56.2 & 38.2 & 20.8 & 39.1 & 48.0 & 37.3 & 59.0 & 40.1 & 34.6 & 55.8 & 36.9 \\
        CAS-ViT-XS~\cite{zhang2024cas} & 12/23 & 162/181 & 36.5 & 56.3 & 38.9 & 21.8 & 39.9 & 48.4 & 37.4 & 59.1 & 40.4 & 34.9 & 56.2 & 37.0 \\
        
        ResNet-50~\cite{he2016deep} & 38/44 & -/260 &36.3 &55.3 &38.6 &19.3 &40.0 &48.8 &38.0 &58.6 &41.4 & 34.4 & 55.1 & 36.7 \\
        FastViT-SA12~\cite{vasu2023fastvit} & -/- & -/- & - & -& -& -& -& - & 38.9 & 60.5 & 42.2 & 35.9 & 57.6 & 38.1 \\
        SHViT-S3~\cite{yun2024shvit} & -/- & -/- & 36.1 &56.6 &38.0 &19.9 &39.1& 50.8& 36.9& 59.4 &39.6& 34.4
        & 56.3 &36.1 \\
        
        \rowcolor{blue!10}\textbf{S2AFormer-T (Ours)} & \textbf{12/23} & \textbf{164/182} & \textbf{36.7} & \textbf{57.0} & \textbf{39.1} & \textbf{21.1} & \textbf{39.7} & \textbf{48.6} & \textbf{37.6} &  \textbf{59.8} & \textbf{40.6} & \textbf{35.4} & \textbf{57.2} & \textbf{37.6} \\ \midrule
        PVT-T~\cite{wang2021pyramid}  & 23/33 & 221/240 & 36.7 & 56.9 & 38.9 & 22.6 & 38.8 & 50.0 & 36.7 & 59.2 & 39.3 & 35.1 & 56.7 & 37.3 \\
        LSNet-S~\cite{wang2025lsnet} &  -/- &  -/- & 36.7 &57.2 &38.6 &20.0 &39.7 & 51.8 &37.4& 59.9 &39.8 &34.8 &56.8 &36.6 \\
        MF-508M~\cite{chen2022mobile} & 8/- & 168/- & 38.0 & 58.3 & 40.3 & 22.9 & 41.2 & 49.7 & - & - & - & - & - & - \\
        EfficientFormer-L1~\cite{li2022efficientformer}  & -/32 & -/196 & - & - & - & - & - & - & 37.9 & 60.3 & 41.0 & 35.4 & 57.3 & 37.3 \\
        
        PVTv2-B0~\cite{wang2021pvtv2} & 13/24 & -/- & 37.2 & 57.2 & 39.5 & 23.1 & 40.4 & 49.7 & 38.2 & 60.5 & 40.7 & 36.2 & 57.8 & 38.6 \\
        
        % CAS-ViT-S & 15/25 & 170/189 & 38.6 & 59.2 &  41.2 & 24.0 & 42.1 & 51.2 & 39.8 & 61.5 & 43.2 & 36.7 & 58.8 & 39.2 \\
        \rowcolor{blue!10}\textbf{S2AFormer-XS (Ours)} & \textbf{13/24} & \textbf{166/185} & \textbf{37.9} & \textbf{58.6} & \textbf{40.3} & \textbf{22.9} & \textbf{41.5} & \textbf{49.8} & \textbf{38.4} &  \textbf{60.2} & \textbf{41.5} & \textbf{35.8} & \textbf{57.3} & \textbf{38.1} \\
        \midrule
        SHViT-S4~\cite{yun2024shvit} & -/- & -/- & 38.8 & 59.8 &41.1 &22.0 &42.4 &52.7 & 39.0 & 61.2 &41.9 & 35.9 & 57.9 & 37.9 \\ 
        EMO-5M~\cite{zhang2023rethinking} & -/- & -/- & 38.9&59.8&41.0&23.8&42.2&51.7 &39.3 &61.7 & 42.4 & 36.4 & 58.4 & 38.7 \\
        ResNet-101~\cite{he2016deep}& 57/63 &315/336 & 38.5 & 57.8 & 41.2 & 21.4 & 42.6 & 51.1 & 40.4 & 61.1 & 44.2 & 36.4 & 57.7 & 38.8 \\
        % \color{blue}SBCFormer-B~\cite{lu2024sbcformer} & \color{blue}-/- & \color{blue}-/- & \color{blue}39.3 &\color{blue}59.8 &\color{blue}41.5 &\color{blue}21.9 &\color{blue}42.7&\color{blue} 53.3 &\color{blue}-&\color{blue}-&\color{blue}-&\color{blue}-&\color{blue}-&\color{blue}- \\
        EfficientViM-M4~\cite{lee2025efficientvim} & -/- & -/- & 38.8 &59.6 &41.1 &22.1 &42.4 &52.8 & 39.3 &60.2 &42.5 &35.8 &57.1 &37.4 \\
        PoolFormer-S24~\cite{yu2022metaformer}  & 31/41  & -/233 & 38.9 & 59.7 & 41.3 & 23.3 & 42.1 & 51.8 & 40.1 & 62.2 & 43.4 & 37.0 & 59.1 & 39.6 \\
        PoolFormer-S36~\cite{yu2022metaformer}  & 41/51 & -/272 & 39.5 & 60.5 & 41.8 & 22.5 & 42.9 & 52.4 & 41.0 & 63.1 & 44.8 & 37.7 & 60.1 & 40.0 \\
        RepViT-M1.1~\cite{wang2024repvit} & -/- & -/- &-&-&-&-&-&- & 39.8 & 61.9 & 43.5 & 37.2 & 58.8 & 40.1 \\
        % ResNeXt101-32x4d~\cite{xie2017aggregated}  & -/63 & -/336 & 39.9 & 59.6 & 42.7 & 22.3 & 44.2 & 52.5 & 41.9 & 62.5 & 45.9 & 37.5 & 59.4 & 40.2 \\
        \rowcolor{blue!10}\textbf{S2AFormer-S (Ours)} & \textbf{17/28} & \textbf{178/197} & \textbf{40.0} & \textbf{60.9} & \textbf{42.7} & \textbf{24.4} & \textbf{43.6} & \textbf{52.9} & \textbf{41.0} &  \textbf{62.5} & \textbf{45.0} & \textbf{37.6} & \textbf{59.7} & \textbf{40.3} \\
        \midrule

        PVT-S~\cite{wang2021pyramid}  & 34/44 & 286/305 & 38.7 & 59.3 & 40.8 & 21.2 & 41.6 & 54.4 & 40.4 & 62.9 & 43.8 & 37.8 & 60.1 & 40.3 \\
        EfficientFormer-L3~\cite{li2022efficientformer} & -/51 & -/250 & - & - & - & - & - & - & 41.4 & 63.9 & 44.7 & 38.1 & 61.0 & 40.4 \\
        SwiftFormer-L1~\cite{shaker2023swiftformer}  & -/31 & -/202 & - & - & - & - & - & - & 41.2 & 63.2 & 44.8 & 38.1 & 60.2 & 40.7 \\
        Swin-T~\cite{liu2021swin}& 38/48 & 248/267 & 41.5 & 62.1 & 44.2 & 25.1 & 44.9 & 55.5 & 42.2 & 64.6 & 46.2 & 39.1 & 61.6 & 42.0 \\
        EfficientFormer-L7~\cite{li2022efficientformer}  & -/101 & -/378 & - & - & - & - & - & - & 42.6 & 65.1 & 46.1 & 39.0 & 62.2 & 41.7 \\
        SwiftFormer-L3~\cite{shaker2023swiftformer}  & -/48 & -/252 & - & - & - & - & - & - & 42.7 & 64.4 & 46.7 & 39.1 & 61.7 & 41.8 \\
        \rowcolor{blue!10}\textbf{S2AFormer-M (Ours)} & \textbf{32/42} & \textbf{234/253} & \textbf{41.7} & \textbf{62.4} & \textbf{44.5} & \textbf{25.8} & \textbf{44.6} & \textbf{55.4} & \textbf{42.6} &  \textbf{64.5} & \textbf{46.9} & \textbf{39.3} & \textbf{62.0} & \textbf{41.7} \\
        \bottomrule
        \label{detection}
        \end{tabular}
    % }
    \end{center}
    \vspace{-2em}
\end{table*}

\section{Experiments}
\label {sect:exp}

In this section, we evaluate the proposed S2AFormer architecture through extensive experiments on large-scale image classification datasets and further assess its adaptability across various downstream vision tasks, including semantic segmentation, object detection, and instance segmentation. 
We first benchmark S2AFormer against existing state-of-the-art approaches, and then conduct ablation studies to identify the key components contributing to its effectiveness.

\subsection{Image Classification}
% \subsubsection{Implementation Details}
We train the proposed S2AFormer models from scratch on ImageNet-1K~\cite{deng2009imagenet} without using pre-trained weights or additional data. Following the strategy outlined in EdgeNeXt~\cite{maaz2022edgenext}, we use a $224 \times 224$ input resolution and train for 300 epochs on four NVIDIA H100 GPUs (96 GB memory). The implementation is done in PyTorch 2.5.1 with Timm 0.4.9, using AdamW~\cite{loshchilov2017decoupled} as the optimizer and a batch size of 1024. The initial learning rate is set to $6\times10^{-3}$ and follows a cosine decay schedule with a 20-epoch warmup. Data augmentation includes label smoothing of 0.1, random resize cropping, horizontal flipping, RandAugment, and a multi-scale sampler. Additionally, we apply EMA~\cite{polyak1992acceleration} with a momentum factor of 0.9995.

% \subsubsection{Results}
The results on the ImageNet-1K~\cite{deng2009imagenet} dataset, shown in Table~\ref{imagenet1k}, highlight the significant improvements achieved by S2AFormer in image classification. 
Compared to state-of-the-art approaches, S2AFormer enhances classification accuracy while effectively balancing model complexity and computational efficiency. 
S2AFormer-mini has 1.28M fewer parameters than that of SHViT-S1~\cite{yun2024shvit} but outperforms it by 2.3\% in accuracy. S2AFormer-T achieves 77.7\% Top-1 accuracy, significantly surpassing models like SHViT-S3~\cite{yun2024shvit}, which has more than twice as many parameters (14.2M) and achieves only 77.4\% accuracy with nearly identical MACs. Similarly, EfficientVMamba-T~\cite{pei2024efficientvmamba}, the top-performing model in the Mamba series, falls short by 1.2\% in accuracy compared to S2AFormer-T, while requiring 0.15G more MACs. S2AFormer-XS outperforms StarNet-S4~\cite{ma2024rewrite}, which has 7.5M parameters, 1.075 GMACs, and 78.4\% accuracy, while being more efficient in terms of parameters and MACs. S2AFormer-S has a slight disadvantage of 0.2\% lower Top-1 accuracy compared to EdgeViT-S~\cite{chen2022edgevit} (81.0\% vs 80.8\%), but it achieves this with much fewer parameters and a reduction of 0.5 GMACs, making it more lightweight. In comparison to InceptionNeXt-T~\cite{yu2024inceptionnext}, which has 28M parameters, 4.2 GMACs, and 82.3\% accuracy, S2AFormer-M is 3.13M smaller in terms of parameter count while delivering the same Top-1 accuracy, further emphasizing its efficiency and reduced computational burden. %\textcolor{blue}{In addition, Fig.~\ref{grad_cam} shows a Grad-CAM~\cite{jacobgilpytorchcam} comparison with the recent LSNet~\cite{wang2025lsnet}, conducted under identical experimental conditions to assess each model’s ability to perceive categories. The results indicate that our method exhibits superior performance in both recognizing and localizing the target categories.}

\subsection{Semantic Segmentation}
% \subsubsection{Implementation Details}
We evaluate the models' performance in semantic segmentation using the ADE20K dataset~\cite{zhou2017scene}, a challenging benchmark for scene parsing. The dataset comprises 20,000 training images and 2,000 validation images, covering 150 detailed semantic categories. The proposed model is tested with Semantic FPN~\cite{kirillov2019panoptic} as the backbone, with normalization layers frozen and pre-trained weights from ImageNet-1K~\cite{deng2009imagenet} classification. Following standard practices~\cite{kirillov2019panoptic,chen2017deeplab}, the network is trained for 40,000 iterations with a batch size of 32, using the AdamW~\cite{loshchilov2017decoupled} optimizer. The initial learning rate is set to 
$2 \times 10 ^ {-4}$ and follows a polynomial decay schedule with a power of 0.9. During training, input images are resized and cropped to $512 \times 512$. The implementation is based on the MMSegmentation~\cite{mmseg2020} framework.

% \subsubsection{Results}
The comparative results %quantitative evaluation on ADE20K~\cite{zhou2017scene}, 
shown in Table~\ref{segmentation} demonstrate that all S2AFormer variants effectively balance computational efficiency and segmentation accuracy. %achieving excellent performance. 
S2AFormer-mini, with just 6M parameters and 23 GFLOPs, achieves a competitive mIoU of 36.7\%, surpassing ResNet-18~\cite{he2016deep} and PVT-T~\cite{wang2021pyramid} by 3.8\% and 1.0\%, respectively. 
With 7M parameters and 25 GFLOPs, S2AFormer-T improves the mIoU to 38.0\%, outperforming PoolFormer-S12~\cite{yu2022metaformer} (16M parameters, 31 GFLOPs, 37.2\% mIoU) and CASViT-XS~\cite{zhang2024cas} (7M parameters, 24 GFLOPs, 37.1\% mIoU) by 0.8\% and 0.9\%, respectively, while maintaining a smaller or comparable model size. S2AFormer-XS achieves 39.2\% mIoU, surpassing EfficientFormer-L1~\cite{li2022efficientformer} by 0.3\% using fewer parameters, demonstrating its high efficiency. 
S2AFormer-S offers better performance than PoolFormer-S24~\cite{yu2022metaformer} (40.8\% vs 40.3\%) with fewer parameters and lower GFLOPs. Similarly, S2AFormer-M outperforms InceptionNeXt-T~\cite{yu2024inceptionnext} (28M parameters, 44 GFLOPs, 43.1\% mIoU) by 0.6\%, while being 7.14\% smaller in parameter count.

\subsection{Object Detection and Instance Segmentation}
% \subsubsection{Implementation Details}
We assess the performance of S2AFormer on object detection and instance segmentation tasks using the COCO2017 dataset~\cite{lin2014microsoft}, which contains 118,000 training images and 5,000 validation images, along with bounding box and mask annotations across 80 categories. We assess the performance of our model with two widely used object detectors: RetinaNet~\cite{ross2017focal} and Mask R-CNN~\cite{he2017mask}. To initialize the backbone, we use weights pre-trained on ImageNet. All models are trained with a batch size of 16 on four H100 GPUs, utilizing the AdamW optimizer~\cite{loshchilov2017decoupled} with an initial learning rate of $1 \times 10 ^ {-4}$. Following standard practices~\cite{ross2017focal,he2017mask,chen2019mmdetection}, we adopt a $1 \times$ (12 epochs) training schedule. The models are implemented using the MMDetection~\cite{chen2019mmdetection} framework.

% \subsubsection{Results}
We conducted a detailed evaluation on the COCO val2017 dataset~\cite{lin2014microsoft}, comparing backbone models based on Average Precision (AP) across various scales (small, medium, large) and tasks (detection and segmentation), along with their efficiency in terms of parameters (Para.) and FLOPs. As shown in Table~\ref{detection}, S2AFormer-mini achieves an impressive 33.4\% $AP$ on RetinaNet~\cite{ross2017focal}, surpassing models like ResNet-18~\cite{he2016deep} and EfficientViT-M4~\cite{liu2023efficientvit} while maintaining fewer parameters and lower GFLOPs. S2AFormer-T achieves 36.7\% $AP$ for RetinaNet~\cite{ross2017focal} and 37.6\% $AP^b$ for Mask R-CNN~\cite{he2017mask}, surpassing PoolFormer-S12~\cite{yu2022metaformer}, which has 36.2\% $AP$ and 37.3\% in $AP^b$, while maintaining a reduced computational burden. Under similar parameter conditions, S2AFormer-XS (with 13/24 parameters and 166/185 GFLOPs) achieves 37.9\% $AP$ for RetinaNet and 38.4\% $AP^b$ for Mask R-CNN, outperforming PVTv2-B0~\cite{wang2021pvtv2} (with 13/24 parameters and 37.2\% $AP$ for RetinaNet and 38.2\% $AP^b$ for Mask R-CNN) by a noticeable margin. Similarly, S2AFormer-M achieves excellent performance across benchmarks.

\begin{figure*}[t]
	\centerline{\includegraphics[width=18cm]{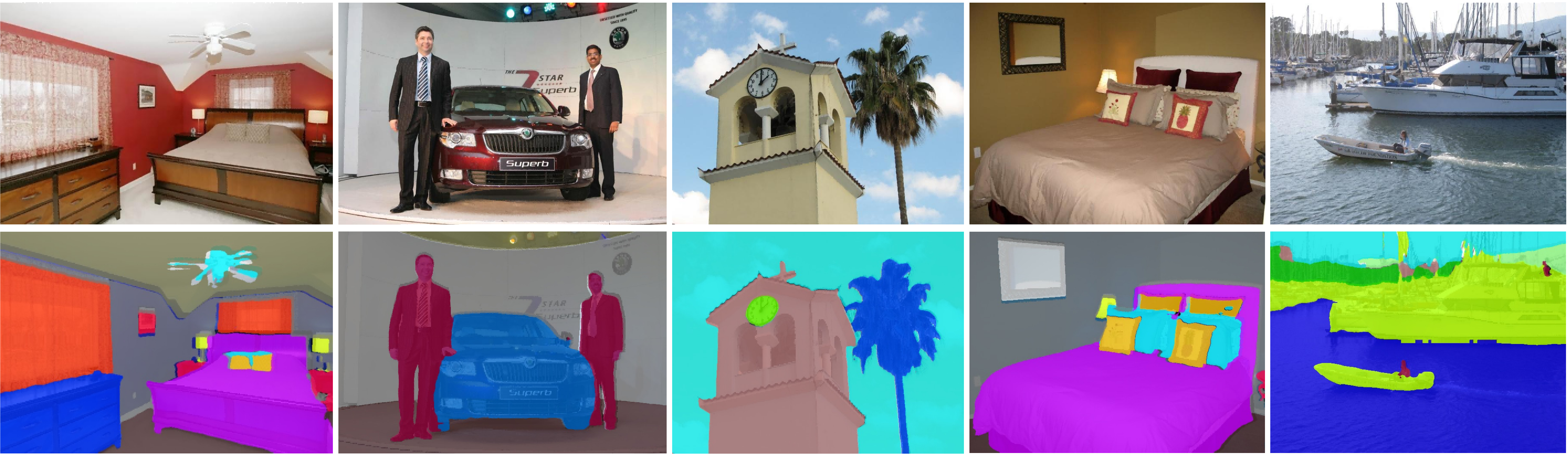}}
	\caption{Qualitative results of semantic segmentation on ADE20K val dataset~\cite{zhou2017scene}. Upper: input validation images, Lower: results generated by S2AFormer-S-based Semantic FPN~\cite{kirillov2019panoptic}.}
	\label{ade_seg}
\end{figure*}

\begin{figure*}[t]
	\centerline{\includegraphics[width=18cm]{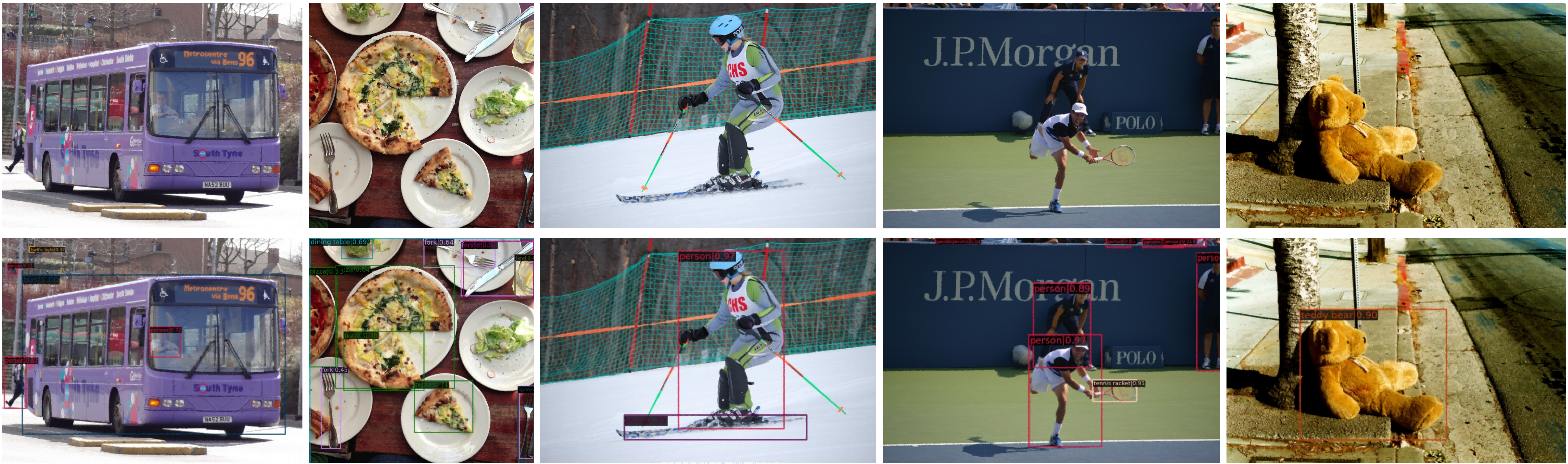}}
	\caption{Qualitative results of object detection on the COCO val2017 dataset~\cite{lin2014microsoft}. Top: input validation images. Bottom: results generated by S2AFormer-S-based RetinaNet~\cite{ross2017focal}.}
	\label{coco_det}
\end{figure*}

\begin{figure*}[t]
	\centerline{\includegraphics[width=18cm]{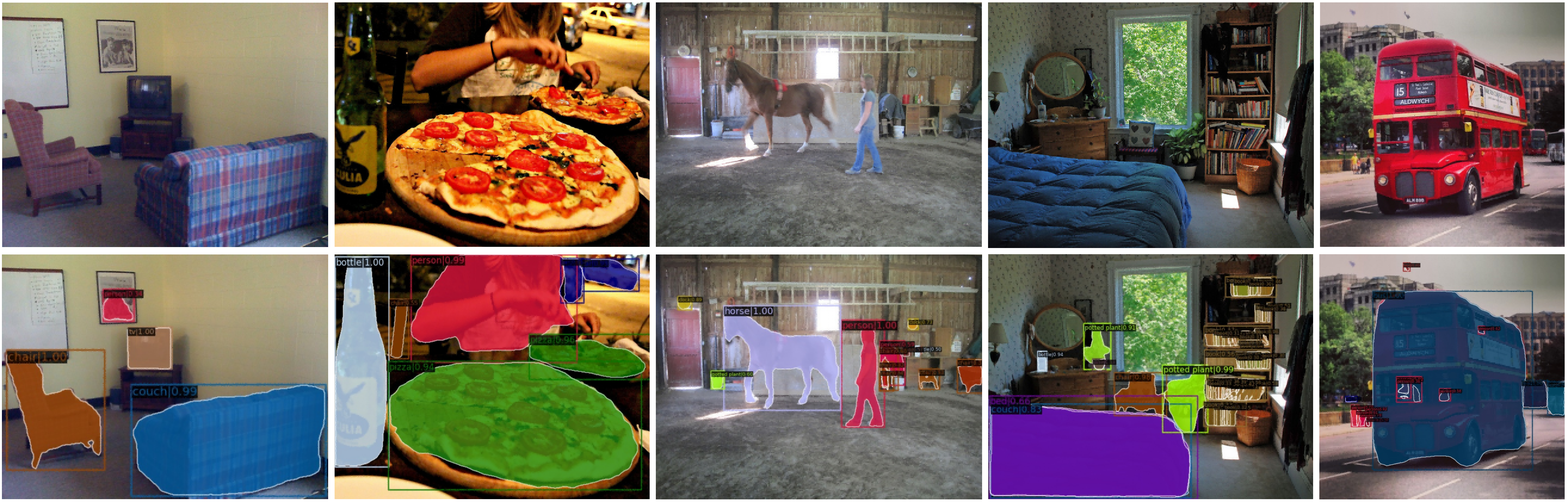}}
	\caption{Qualitative results of instance segmentation on the COCO val2017 dataset~\cite{lin2014microsoft}. Top: input validation images. Bottom: results generated by S2AFormer-S-based Mask R-CNN~\cite{he2017mask}.}
	\label{coco_ins}
\end{figure*}

\subsection{Qualitative Results}
The qualitative results in Fig.~\ref{ade_seg} highlight the efficacy of our method in generating visually coherent and fine-grained segmentation maps on the ADE20K validation set~\cite{zhou2017scene}. Complementing these findings, Figs.~\ref{coco_det} and~\ref{coco_ins} showcase task-specific visualizations from our S2AFormer-S model, addressing instance segmentation and object detection challenges, respectively. 
These outcomes underscore the model’s robustness in accurately localizing and delineating objects across diverse and cluttered scenes, even under varying scales and occlusions.

\begin{table}[t]
\caption{Throughput comparison across GPU, CPU, and ONNX platforms. GPU: NVIDIA H100 NVL, CPU: Intel(R) Xeon(R) Gold 6426Y.}
% \vspace{-2em}
\label{throughput}
\begin{center}
\scalebox{0.9}{
\begin{tabular}{l|ccccc}
\toprule
\multicolumn{1}{l|}{\multirow{2}{*}{Model}} &\multicolumn{1}{c}{\multirow{2}{*}{\#P.$\downarrow$}} &\multicolumn{1}{c}{\multirow{2}{*}{Top-1$\uparrow$}}& \multicolumn{3}{c}{Throughput (images/s)}          \\
\multicolumn{1}{c|}{} &&              & \multicolumn{1}{c}{GPU$\uparrow$} & \multicolumn{1}{c}{CPU$\uparrow$} & \multicolumn{1}{c}{$\mathrm{ONNX}$$\uparrow$} \\ \midrule
\midrule
% EfficientViT-M0~\cite{liu2023efficientvit}&2.3&63.2&8735&&\\
% EfficientViT-M1~\cite{liu2023efficientvit}&3.0&68.4&7963&&\\
% PVTv2-B0~\cite{wang2021pvtv2}&3.7&70.5&7662&&\\
% EfficientViT-M2~\cite{liu2023efficientvit}&4.2&70.8&6855&&\\
EMO-1M~\cite{zhang2023rethinking}&1.3&71.5&4674&45.57&118.72\\
EfficientViT-M3~\cite{liu2023efficientvit}&6.9&73.4&4355&35.34&169.35\\
MobileMamba-T2~\cite{he2024mobilemamba} & 8.8 & 73.6 & 6279 & 63.75 & 178.38\\
EfficientViT-M4~\cite{liu2023efficientvit}&8.8&74.3&4289&34.34&159.51\\
% EfficientFormer-L7~\cite{li2022efficientformer}&82.1&83.3&686&&\\
EMOv2-1M~\cite{zhang2025emov2} & 1.4 & 72.3 &  4886&  45.70&  112.98\\
EMO-2M~\cite{zhang2023rethinking}&2.3&75.1&3359&41.20&83.79\\
PVT-T~\cite{wang2021pyramid}&13.2&75.1&5033&61.01&117.63\\
\rowcolor{blue!10} \textbf{S2AFormer-mini} & \textbf{5.0}  & \textbf{75.1}   & \textbf{6330} & \textbf{58.21}& \textbf{118.26} \\ \midrule

FastViT-T8~\cite{vasu2023fastvit}&3.6&75.6&7615&50.76&129.27\\
EfficientFormerV2-S0~\cite{li2023rethinking}&3.5&75.7&520&44.57&136.32\\
MobileMamba-T4~\cite{he2024mobilemamba} & 14.2 & 76.1 & 5124& 52.33 & 146.37\\
EdgeViT-XXS~\cite{chen2022edgevit} &4.1 &74.4 &3614 &50.82&134.79\\
 PoolFormer-S12~\cite{yu2024metaformer} & 12.0 & 77.2 &2148 & 32.67& 102.16\\
EfficientViT-M5~\cite{liu2023efficientvit}&12.4&77.1&3843&28.48&126.28\\
EdgeViT-XS~\cite{chen2022edgevit} &6.7&77.5&2872&35.14&106.10\\
\rowcolor{blue!10} \textbf{S2AFormer-T} & \textbf{5.8}  & \textbf{77.7}   & \textbf{4768} &\textbf{41.75} & \textbf{112.89} \\ \midrule
% \rowcolor{blue!10} \textbf{S2AFormer-T} & \textbf{-}  & \textbf{-}   & \textbf{4739/2530A100} && \\ \midrule

EMO-5M~\cite{zhang2023rethinking}&5.1&78.4&2370&38.18&62.52\\
PVTv2-B1~\cite{wang2021pvtv2}&14.0&78.7&3930&79.51&37.96\\
EMO-6M~\cite{zhang2023rethinking}&6.1&79.0&2102&35.90&57.00\\
StarNet-S4~\cite{ma2024rewrite} & 7.5 & 78.4 & 2570& 30.83&64.26\\
EfficientFormerV2S1~\cite{li2023rethinking}&6.1&79.0&472&37.89&104.35\\
EfficientFormer-L1~\cite{li2022efficientformer}&12.3&79.2&3572&66.93&88.72\\
iFormer-S~\cite{zheng2025iformer} & 6.5 & 78.8 &  3764&  32.89&  66.30\\
\rowcolor{blue!10} \textbf{S2AFormer-XS} & \textbf{6.5}  & \textbf{78.9}   & \textbf{4185} &\textbf{34.09}&\textbf{71.63} \\  \midrule

% RepViT-M0.9~\cite{wang2024repvit} &5.1&78.7&6068&&\\
% RepViT-M1.1~\cite{wang2024repvit} &8.2&80.7&5092&&\\
% RepViT-M2.3~\cite{wang2024repvit} &14.0&82.3&2978&&\\

FastViT-T12~\cite{vasu2023fastvit}&6.8&79.1&5277&42.06&88.31\\
FastViT-S12~\cite{vasu2023fastvit}&8.8&79.8&4952&39.41&68.09\\
FastViT-SA12~\cite{vasu2023fastvit}&10.9&80.6&4812&38.48&50.12\\
% FastViT-SA24~\cite{vasu2023fastvit}&20.6&82.6&2749&&\\
LSNet-B~\cite{wang2025lsnet} & 23.2 & 80.3 & 2890&27.20 & 52.88\\
PVT-S~\cite{wang2021pyramid}&24.5&79.8&2736&43.50&53.01\\
EdgeViT-S~\cite{chen2022edgevit} &11.1&81.0&1951&23.15&58.32\\
EfficientViT-M5~\cite{liu2023efficientvit} & 12.4 & 80.8 & 2396&29.05&54.62\\
\rowcolor{blue!10} \textbf{S2AFormer-S} & \textbf{10.7}  & \textbf{80.8}   & \textbf{2644} &\textbf{25.78} & \textbf{55.84} \\  \midrule

PVT-M~\cite{wang2021pyramid}&44.2&81.2&1717&25.96&36.89\\
EfficientFormerV2-S2~\cite{li2023rethinking}&12.6&81.6&256&25.08&46.94\\
PVT-L~\cite{wang2021pyramid}&61.4&81.7&587&18.84&27.77\\
% \color{blue}InceptionNeXt~\cite{yu2024inceptionnext} &\color{blue}28.0 & \color{blue}82.3 &\color{blue} 2359&\color{blue}30.50&\color{blue}34.12\\
EfficientVMamba-B~\cite{pei2024efficientvmamba} & 33.0 & 81.8 & 2432& 26.75& 32.54\\
PVTv2-B2~\cite{wang2021pvtv2}&25.4&82.0&2218&40.20&40.83\\
EfficientFormer-L3~\cite{li2022efficientformer}&31.3&82.4&1550&35.96&25.31\\
VRWKV-B~\cite{duan2024vrwkv} & 93.7 & 82.0 &  1520& 20.49 &  24.68\\
\rowcolor{blue!10} \textbf{S2AFormer-M} & \textbf{24.9}  & \textbf{82.3}   & \textbf{1652}&\textbf{15.50}& \textbf{27.15} \\  \midrule

\end{tabular}
}

\vspace{-2em}
\end{center}
\end{table}

\subsection{Inference Speed}

% We implemented S2AFormer across three distinct platforms—GPU, CPU, and ONNX—and evaluated its efficiency through throughput analysis. 
% Table~\ref{throughput} compares throughput across these platforms, tested on an NVIDIA H100 NVL GPU and an Intel(R) Xeon(R) Gold 6426Y CPU. 
We implemented S2AFormer on three different platforms and evaluated its efficiency through throughput analysis. Note that, to ensure fairness and reproducibility, we report GPU, CPU, and ONNX throughput using a single codebase and the same checkpoint, rather than referencing results from prior works. All experiments were conducted on an NVIDIA H100 NVL GPU and an Intel Xeon Gold 6426Y CPU, using PyTorch 2.5.1 (for GPU/CPU) and ONNX Runtime (for ONNX).

As shown in Table~\ref{throughput}, for GPU-based inference, S2AFormer demonstrates near-optimal performance with comparable ImageNet-1K accuracy. For example, S2AFormer-mini achieves a throughput of 6330 images/s on the GPU, significantly outperforming other models. It exceeds the throughput of EMO-2M~\cite{zhang2023rethinking}, which achieves 3359 images/s, making S2AFormer-mini nearly twice as fast in GPU throughput. 
%The GPU efficiency of S2AFormer-mini 
This highlights its strong image processing capabilities compared to similar models like EfficientViT-M3~\cite{liu2023efficientvit} (4355 images/s) and PVT-T~\cite{wang2021pyramid} (5033 images/s). 

In CPU and ONNX-based inference, S2AFormer also shows highly competitive performance. 
For example, S2AFormer-T surpasses EdgeViT-XS~\cite{chen2022edgevit} in throughput on both CPU (41.75 images/s) and ONNX (112.89 images/s) platforms. 
This highlights the model’s robustness in non-GPU environments, where inference speed can become a bottleneck. This is particularly important for real-world deployment scenarios, where model efficiency across various hardware platforms, including CPUs and optimized ONNX models, plays a crucial role in practical applications.

\subsection{Ablation Studies}

Our comparative experiments across three vision benchmarks demonstrate the robustness and efficiency of the proposed strip self-attention in S2AFormer. 
In addition to the core HPBs, we introduced a specialized LIM to compensate for the limited local perceptual capacity of self-attention. 
In this section, we conduct ablation studies to validate the effectiveness of both LIM and the convolution-based spatial reduction operation.

\begin{table}[t]
\caption{Performance comparisons with and without LIM on COCO val2017 dataset~\cite{lin2014microsoft}.}
\begin{center}
\scalebox{0.9}{
\begin{tabular}{l|c|c|ccc}
\toprule
\multicolumn{1}{l|}{\multirow{2}{*}{Model}} &\multicolumn{1}{c|}{\multirow{2}{*}{LIM}} & \multicolumn{1}{c|}{\multirow{2}{*}{Top-1$\uparrow$}} & \multicolumn{3}{c}{RetinaNet 1$\times$}          \\
&&\multicolumn{1}{c|}{}                          & \multicolumn{1}{c}{\#Para.(M)$\downarrow$} & \multicolumn{1}{c}{GFLOPs$\downarrow$} & \multicolumn{1}{c}{AP(\%)$\uparrow$} \\ \midrule
\midrule
\multicolumn{1}{l|}{\multirow{2}{*}{S2AFormer-mini}}&\XSolidBrush& 74.89\% & 11.55 & 159.06 & 32.6\\
&\Checkmark & 75.06\%& 11.65 & 159.20 & 33.4\\
\midrule
\multicolumn{1}{l|}{\multirow{2}{*}{S2AFormer-T}}&\XSolidBrush& 77.63\% & 12.29 & 163.50 & 35.3\\
&\Checkmark & 77.73\% & 12.44 & 163.71 & 36.7\\
\bottomrule
\end{tabular}
\label{ablation_LIM}
}
\end{center}
\end{table}

\begin{table}[t]
\caption{Performance comparisons between convolution and pooling operations for spatial reduction purposes using Top-1 accuracy and inference throughput.}
\begin{center}
\scalebox{0.9}{
\begin{tabular}{l|ccc|cc}
\toprule
\multicolumn{1}{l|}{\multirow{2}{*}{Model}}&\multicolumn{1}{c}{\multirow{2}{*}{\#Para.(M)$\downarrow$}}&\multicolumn{1}{c}{\multirow{2}{*}{GMACs$\downarrow$}} & \multicolumn{1}{c|}{\multirow{2}{*}{Top-1$\uparrow$}} & \multicolumn{2}{c}{TP (imgs/s)}          \\
&&&\multicolumn{1}{c|}{}                          & \multicolumn{1}{c}{GPU$\uparrow$} & \multicolumn{1}{c}{CPU$\uparrow$}  \\ \midrule
\midrule
\textbf{S2AFormer-mini}& 5.02 & 0.43 & 75.1\% & 6330 & 58.21\\

 \multicolumn{1}{r|}{Conv $\rightarrow$ Pooling}& 5.01 & 0.57 & 75.0\% & 3649 &54.10\\
\midrule
\textbf{S2AFormer-T}& 5.80 & 0.66 & 77.7\% & 4768 & 41.75\\

Conv $\rightarrow$ Pooling& 5.79 & 1.08 & 77.3\% & 1695 & 34.20\\

\bottomrule
\end{tabular}
\label{ablation_pooling}
}
\end{center}
\end{table}

% \subsubsection{Comparison between SSA and SA}

\subsubsection{Effectiveness of LIM} 
As shown in Table~\ref{ablation_LIM}, for S2AFormer-Mini, the parameter count increases slightly from 11.55M to 11.65M, while the GFLOPs remain nearly unchanged, with values of 159.06 and 159.20, respectively. Initially, the model achieves a Top-1 accuracy of 74.89\%. However, after incorporating LIM, there is a noticeable improvement in performance, with the Top-1 accuracy rising to 75.06\%. 
A similar trend is observed for S2AFormer-T. Without LIM, the model achieves a Top-1 accuracy of 77.63\%, but with LIM, the accuracy increases slightly to 77.73\%. The parameter count for this model increases modestly from 12.29M to 12.44M, and GFLOPs show a minor increase from 163.50 to 163.71. Despite these small increases in computational cost, the model's Average Precision (AP) improves significantly, with LIM boosting the AP from 35.3 to 36.7. This demonstrates that the improvements in accuracy come with minimal additional computational overhead, particularly in terms of detection performance.

\subsubsection{Effectiveness of Conv Spatial Reduction} 
As shown in Table~\ref{ablation_pooling}, our comparison between Convolution and Pooling for reducing spatial dimensions reveals that Convolution slightly outperforms Pooling in both the S2AFormer-mini and S2AFormer-T models. 
Specifically, S2AFormer-mini achieves a Top-1 accuracy of 75.1\% with Convolution, which is slightly higher than the 75.0\% obtained with Pooling. Similarly, S2AFormer-T reaches a Top-1 accuracy of 77.7\% using Convolution, compared to 77.3\% with Pooling.

In terms of efficiency, Convolution also demonstrates superiority, requiring fewer MACs: 0.43 GMACs versus 0.57 GMACs for S2AFormer-mini and 0.66 GMACs versus 1.08 GMACs for S2AFormer-T. Regarding inference speed, Convolution outpaces Pooling by a significant margin. On the GPU, S2AFormer-mini with Convolution processes 6330 images/s, compared to 3649 images/s with Pooling. Similarly, S2AFormer-T processes 4768 images/s with Convolution, versus 1695 images/s with Pooling. On the CPU, Convolution also leads, with 58.21 images/s for S2AFormer-mini and 41.75 images/s for S2AFormer-T, while Pooling reaches 54.10 images/s and 34.20 images/s, respectively.

In conclusion, convolution consistently delivers better accuracy, lower computational complexity, and faster inference speeds on both GPU and CPU, making it the more efficient option for spatial dimension reduction in our model.

\subsection{Limitations and Future Works}
While our proposed S2AFormer strikes an effective balance between efficiency and accuracy, there is still potential to refine the spatial–channel compression strategy. In future work, we aim to explore learnable token compression mechanisms that dynamically adapt to the content distribution in both spatial and channel dimensions. A promising approach is to integrate mask-guided autoregressive compression, inspired by MAE~\cite{he2022masked}, to selectively retain informative tokens for global reasoning while discarding redundant ones. This adaptive compression could further reduce computational costs without sacrificing accuracy, enhancing the scalability and robustness of our framework for more complex dense prediction tasks.

\section{Conclusion}
\label {sect:conclusion}

In this work, we developed a new family of hybrid architecture that efficiently combines CNNs and Vision Transformers, called S2AFormer. Our approach reduces redundancies in both the spatial and channel dimensions, contributing to a more efficient architectural design. Extensive experiments across three downstream benchmarks have validated the effectiveness and superiority of S2AFormer over some state-of-the-art methods. The results highlight S2AFormer as a promising solution for achieving both high performance and efficiency in computer vision tasks.

%\newpage

\bibliographystyle{IEEEtran}
\bibliography{reference}

\vfill

\end{document}